\PassOptionsToPackage{usenames,dvipsnames}{xcolor}
\documentclass[runningheads]{llncs}
\usepackage[T1]{fontenc}
\usepackage{graphicx}
\usepackage{xcolor}
\usepackage[nofillcomment,ruled,linesnumbered,norelsize,noend]{algorithm2e}
\usepackage{amssymb}
\usepackage{amsmath}
\usepackage{enumitem}

\usepackage{cite} 

\usepackage{etoolbox}  
\newtoggle{arxiv}  
\toggletrue{arxiv}

\renewcommand{\paragraph}[1]{\smallskip\noindent\textbf{#1}~}

\newcommand{\restset}{R}
\newcommand{\incset}{\Gamma}

\newcommand{\QED}{\hfill\rule{2mm}{2mm}}

\newcommand{\dme}{\mathcal{E}}
\newcommand{\dmo}{\mathcal{O}}
\newcommand{\dmu}{\mathcal{U}}

\DeclareMathOperator*{\argmax}{arg\,max}
\DeclareMathOperator*{\argmin}{arg\,min}
\DeclareMathOperator{\cost}{\rho} 
\DeclareMathOperator*{\benefit}{g} 

\newcommand{\agset}{{\mbox{$\mathbb{X}$}}} 

\newcommand{\rset}{{\mbox{$\mathbb{Y}$}}} 
\newcommand{\srset}{\mathbb{Y}}  
\newcommand{\sag}{x^*} 
\newcommand{\sagrset}{\rset_I} 

\newcommand{\cnp}{\mbox{\textbf{NP}}}
\newcommand{\cnump}{\mbox{\textbf{\#P}}}

\newcommand{\prin}{\textsc{Principal}}
\newcommand{\budget}{\beta}

\newcommand{\tproblem}{\textsc{MatchingAdvice}}
\newcommand{\tdproblem}{\textsc{D-MatAdv}}

\newcommand{\maxcover}{\mbox{\textsc{Max-Coverage}}}
\newcommand{\xbenefit}{\mbox{$\psi$}}

\newcommand{\scmr}{\text{Single-Choice-Multi-Restriction}}
\newcommand{\mcsr}{\text{Multi-Choice-Single-Restriction}}
\newcommand{\scsr}{\text{Single-Choice-Single-Restriction}}
\newcommand{\mcmr}{\text{Multi-Choice-Multi-Restriction}}

\newcommand{\scmrs}{\text{Single-C-Multi-R}}
\newcommand{\mcsrs}{\text{Multi-C-Single-R}}
\newcommand{\mcmrs}{\text{Multi-C-Multi-R}}
\usepackage{pdfpages}

\begin{document}
%
%
\title{Resource Allocation to Agents with Restrictions:
Maximizing Likelihood with Minimum Compromise}

\titlerunning{Resource Allocation with Restrictions}
%
\author{Yohai Trabelsi\inst{1}
\and
Abhijin Adiga\inst{2} 
\and
Sarit Kraus\inst{1} 
\and
S.~S. Ravi\inst{2,3}
}
\institute{Department of Computer Science, 
Bar-Ilan University, Ramat Gan, Israel
\email{yohai.trabelsi@gmail.com,
sarit@cs.biu.ac.il}
\and
Biocomplexity Institute and Initiative,
Univ. of Virginia, Charlottesville, VA, USA\newline
\email{abhijin@virginia.edu}
\and
Dept. of Computer Science, University at Albany -- SUNY, Albany, NY, USA\newline
\email{ssravi0@gmail.com}
}
%
%

%
\maketitle              
\begin{abstract}
Many scenarios where agents with restrictions compete for resources can be
cast as maximum matching problems on bipartite graphs. 
Our focus is on resource allocation problems where agents may have restrictions that make them incompatible with some resources.
We assume that a \prin{} chooses a maximum
matching randomly so that each agent 
is matched to a resource with some probability.
Agents would like to improve their
chances of being matched by modifying their restrictions within certain limits.  
The \prin{}'s goal is to advise an unsatisfied agent
to relax its restrictions 
so that the total cost of relaxation is within a budget
(chosen by the agent) and the increase in the probability of being assigned a resource is maximized.
We establish hardness results for some variants of
this budget-constrained maximization problem and present 
algorithmic results for other variants. 
We experimentally evaluate our
methods on synthetic datasets as well as on two novel 
real-world datasets: a
vacation activities dataset and a classrooms dataset.



\end{abstract}

\section{Introduction}
\label{sec:intro}

There are many practical contexts where a set of \textbf{agents} 
must be suitably matched with a set of \textbf{resources}.
Examples of such contexts include matching classes with classrooms~\cite{Phillips2015integer},
medical students with hospitals \cite{roth1986allocation},
matching buyers with products \cite{lu2009role}, 
matching customers with taxicabs~\cite{ghoseiri2010real}, 
matching agricultural equipment 
with farms~\cite{Gilbert-2018,RS-2020}, etc.
We assume that the matching process assigns at most one resource to 
each agent and that each resource is assigned to at most one agent. 
It is possible that some agents are not assigned resources
and some resources are unused.


Agents have \textbf{restrictions} (or {preferences}) while resources
have \textbf{constraints}. We assume that agents' restrictions
are \emph{soft}; that is, agents are willing
to \emph{relax} their restrictions so that they can get
a resource. 
An agent who is unwilling to compromise may not
get any resource.
However, the constraints associated with resources are \emph{hard}; they \emph{cannot} be relaxed.

\noindent
\textbf{Example:}~ An instructor who indicates her restriction for 
the classroom capacity as 
``Capacity $\geq$ 70" may be willing to relax this restriction to 
``Capacity $\geq$ 60" to improve her chances of obtaining
a classroom.
However, a classroom of size 50  
imposes the hard constraint ``Capacity $\leq$ 50".

An agent  is \textbf{compatible} with a resource 
(i.e., the agent can be matched with the resource)
only when the (hard) constraints of the resource are satisfied
by the agent's restrictions. 
The problem of assigning resources to agents can be modeled
as a matching problem on the following bipartite graph,
which we refer to as the \textbf{compatibility graph}: the graph
has two disjoint sets
of nodes corresponding to the agents and resources respectively; 
each edge $\{u,v\}$ in the graph indicates that the agent represented
by $u$ is compatible with the resource represented by $v$. 
A \prin{} (who is not one of the agents)
chooses a maximum matching in the graph to maximize the number of agents
who are assigned resources. Usually, there are many such maximum matchings,
each one allocating resources to a (possibly) different set of agents.
For fairness, the \prin{} chooses a maximum matching randomly out of a given distribution. 
The \prin{} may use, for example, an algorithm for fair matching~\cite{garcia2020fair} or a straight-forward process that randomly orders the agents and uses a deterministic matching algorithm like the Hopcroft-Karp algorithm \cite{Hopcroft-Karp-1973} to generate a maximum matching. 

It is natural for an agent, who is concerned that she will not be matched in the randomly generated matching, to seek advice from the \prin{}
in the form of changes to her restrictions 
in order to increase the likelihood of getting matched.
We assume a nonnegative cost associated with relaxing each restriction.
Agents are desperate to get such advice when there are several rounds of matching and they failed in previous ones; such a situation arises, for example,  in the case of medical students who were not matched during the first round of the residency matching process \cite{Izenberg18}.
Developing such recommendations 
can be modeled as the following budget-constrained 
optimization problem: 
find a set of modifications to an unmatched agent's restrictions
under a budget constraint so that the likelihood of the agent being
matched to a resource is maximized, given the resource compatibility information for 
the other agents.

Several recommendation systems in environments where agents compete for resources are similar to our notion of a \prin. As an example, many
route planning and satellite navigation apps provide advice to a given agent (driver) without taking into account possible changes in the behaviors of other agents due to similar recommendations. These recommendations often lead to undesirable consequences that are referred to as the price of anarchy \cite{wapner2020gps}. The study of how to decrease the price of anarchy is beyond the scope of this paper. 

\paragraph{Summary of contributions.}

\smallskip
\noindent
\underline{1. The matching advice problem.}~
We develop a formal framework for advising agents in a resource 
allocation setting viewed as a matching problem on an
agent-resource bipartite graph.
We formulate a budget-constrained
optimization problem to generate suitable relaxations of an unmatched
agent's restrictions so as to maximally increase the 
probability that the agent will be matched. We identify and study
different forms of restrictions arising from agent restrictions
and resource properties in real-world applications.

\noindent
\underline{2. Complexity of improving the 
likelihood of matching.}
We show that, in general, the budget-constrained optimization problem is \cnp-hard.

\noindent
\underline{3. Algorithms for improving the likelihood of matching.}
Under uniform costs for relaxing restrictions and
uniform random selection of maximum matchings,
we present algorithmic results for some classes of restrictions 
(which will be defined in Section~\ref{sse:incompat_type}).
Specifically, we present an 
efficient approximation algorithm (with
a performance guarantee of $(1-1/e)$) for the Multi-Choice Single-Restriction case.
This result relies on the submodularity of 
the objective function. 
For another class called \emph{threshold-like} restrictions, we develop a fixed parameter tractable algorithm, assuming
that the budget and the cost of removing each restriction are non-negative integers.

\noindent
\underline{4. Experimental Study.}~ We study the performance
of our recommendation algorithms on both synthetic data sets
as well as two real-world data sets.
The latter data sets arise in the contexts of assigning classrooms to
courses and matching children with activities. We evaluate our algorithms 
under different cost schemes. The insights gained from this study
can inform the \prin{} (e.g., university administration) on 
issues such as adding, removing or modifying resources to cater to
the needs of agents.

\paragraph{Related work.}
Resource allocation in multi-agent systems has been studied 
by a number of researchers (e.g., \cite{Chevaleyre-etal-2006,Gorodetski-etal-2003,Dolgov-etal-2006}).
The general focus of this work is on topics such as how agents express their resource requirements, algorithms for allocating resources to satisfy those requirements and evaluating the quality of the resulting
allocations.
Nguyen~et~al.~\cite{nguyen2013survey} discuss some complexity and approximability in this context~\cite{nguyen2013survey}.
Zahedi et al. \cite{Zahedi-etal-2020} study the problem of allocating tasks to agents in such a way that the task allocator
can respond to queries dealing with counterfactual allocations. 

Motivated by e-commerce applications, Zanker et al. \cite{Zanker-etal-2010} discuss the design and evaluation of constraint-based recommendation systems that allow users to specify soft constraints regarding products of interest.
These constraints are in the form of rank ordering of
desired products.
Both algorithms for the problem and a system  which includes implementations of those algorithms
are discussed in \cite{Zanker-etal-2010}.
Felfernig~et~al.~\cite{Felfernig-etal-2011} provide a 
discussion on the design of constraint-based recommendation
systems and the technologies that are useful in developing such
systems.
Parameswaran et al. \cite{Parameswaran-etal-2011} discuss the development of a recommendation system that allows university students to choose courses; the system has the capability to handle complex constraints specified by students as
well as those imposed by courses.
Zhou and Han \cite{Zhou-Han-2019} propose an approach
for a graph-based recommendation system that groups together
agents with similar restrictions to allocate resources.
To our knowledge, the problem studied in our paper, namely advising agents
to modify their restrictions to improve their chances of obtaining resources,  has not been addressed in the literature.

\iftoggle{arxiv}{
\noindent
\textbf{Note:}~ Proofs of many results mentioned in the paper appear in the appendix.
}{
\noindent
\textbf{Note:}~ For space reasons, most of the proofs do
not appear in this version; they can be found in \cite{TA+2022}.
}

\newcommand{\attribs}{\mathcal{A}}
\newcommand{\boolattribs}{\mathcal{B}}

\section{The Matching Advice Framework}
\label{sec:definitions}

\subsection{Graph Representation and Problem Formulation}
\noindent
\textbf{Agents, resources, and compatibility.} We consider
scenarios consisting of a set of \emph{agents} (denoted by \agset) and a set of
\emph{resources} (denoted by \rset). Every agent would like to be matched
to a resource.  However, agents may have \emph{restrictions}
 that prevent them from being matched to certain
resources. Such agent-resource pairs are said to be \emph{incompatible}.
We represent this agent-resource relationship using an
$\agset\rset$-bipartite graph called the \emph{compatibility
graph}~$G(\agset,\rset,E)$, where the edge~$\{x,y\}\in E(G)$ iff the agent $x \in \agset$ is
compatible with $y\in\rset$.
A \prin{} assigns resources to agents. To maximize resource usage, the \prin{} picks a \emph{maximum matching}~\cite{CLRS-2009}
from the compatibility graph.

\paragraph{The advice seeking agent and its restrictions.}
The special
agent who seeks advice will henceforth be denoted by~$\sag$.
Let~$\sagrset\subseteq\rset$ be the set of resources that are incompatible
with~$\sag$. 
Let~$\restset=\{r_1,r_2,\ldots,r_\ell\}$ be the set of restrictions
of~$\sag$.
A \emph{resource}--\emph{restrictions} pair~$(y,\restset')$ consists of
a resource~$y$ and a restriction set~$\restset'\subseteq\restset$ such
that (i)~$y$ is incompatible with~$\sag$ and (ii)~$\restset'$ is a minimal
set of restrictions to remove so that~$y$ becomes compatible with~$\sag$.
A resource-restriction pair
describes precisely why resource $y$ is currently incompatible with 
$\sag$ (i.e., the edge $\{\sag, y\}$ is not in the compatibility graph), and how it can be made compatible. Suppose a set $A$ of restrictions is removed. Then, 
a previously incompatible resource $y$ becomes compatible iff there exists
$(y,R')\in \Gamma$ such that $R'\subseteq A$. We then add the new edge $\{\sag, y\}$
to the compatibility graph.
Let~$\incset=\{(y,\restset')\mid y\in\rset,\, \restset'\subseteq\restset\}$
be the set of such resource--restrictions pairs. We refer to $\Gamma$ as the
\emph{incompatibility set} of $\sag$. Note that there could be more than one
resource--restrictions pair with the same resource when there are 
multiple choices for removing restrictions to make the resource compatible with $\sag$. For a restriction~$r\in\restset$, let~$\cost(r)$ be a
positive real number denoting the cost incurred by~$\sag$ for relaxing~$r$.
For any~$A\subseteq\restset$, the cost of relaxing all the restrictions in~$A$
is~$\cost(A)=\sum_{r\in A}\cost(r)$.

\paragraph{Resource allocation using bipartite maximum matching.} 
To maximize resource usage and ensure fairness for all
agents, we assume that the \prin{} picks a maximum matching from the
set of all possible maximum matchings. There are two components to this
part of the framework: (i)~generating a random maximum matching of the
compatibility graph and (ii)~computing the probability that~$\sag$ is
picked in a random maximum matching. Firstly, we note that a maximum
matching of a bipartite graph can be obtained in polynomial
time~\cite{Hopcroft-Karp-1973}. Given any deterministic algorithm for
maximum matching, one can permute the set of agents or resources (or both)
randomly and obtain a random matching or one could use approaches such as
the fair matching algorithm~\cite{garcia2020fair}.  In any case, the first part can be computed in polynomial time. 
The second part however is computationally harder.  The
distribution from which the matching is sampled depends on the algorithm
used by the \prin{}. In order to provide advice to an agent, the \prin{}
must find the probability that a maximum matching chosen from this
distribution includes that agent.  This problem is closely related to a
computationally intractable (technically, \cnump-hard) problem, namely
counting the number of maximum matchings in bipartite graphs (or sampling
them uniformly)~\cite{valiant1979complexity,Jerrum-1987}.

One way to estimate this
probability is as follows: given an algorithm that generates a random maximum matching,
sample a large number of maximum matchings and compute the ratio of the
number of matchings in which~$\sag$ was matched to the total number of
samples.

\paragraph{The advice framework.}
The following are the steps in the maximum matching advice framework, given
the set of agents~$\agset$ and the set of resources~$\rset$.
\begin{enumerate}[leftmargin=*,noitemsep]
\item An agent~$\sag$ approaches the \prin{} seeking advice. The inputs to
the framework are a compatibility graph~$G$, the restrictions
set~$\restset$ of~$\sag$, and its incompatibility set~$\incset$.
\item The \prin{} computes (or estimates) the probability that~$\sag$ is matched to a
resource and provides this information
to $\sag$.
\item If~$\sag$ is not satisfied with the probability, then, it specifies the
cost~$\cost(\cdot)$ of relaxing its restrictions and a budget~$\budget$ as an
upper bound for the cost it is willing to pay.
\item The \prin{} suggests a relaxation solution (if one exists)  that
results in an \emph{augmented compatibility graph}~$G'$ for which the
improvement in probability of the agent being matched is maximized under the
budget constraint.
\end{enumerate}

\paragraph{The Probability Gain.} 
Let us denote $G$ as the original compatibility graph and $G'$ as the new compatibility graph obtained by adding edges after relaxing the restrictions~$\restset^*$
chosen by the special agent $x^*$.
Denote by~$p(G)$ and~$p(G')$ the probability that~$x^*$ is matched in
a maximum matching of~$G$ and~$G'$ respectively.
The probability gain $g(\restset^*)$ is defined as $p(G') - p(G)$.
Since $p(G)$ does not change when $x^*$ relaxes some
restrictions, maximizing $g(\restset^*)$ is equivalent to maximizing $p(G')$.
Now, we define the \tproblem{} problem formally.

\smallskip
\noindent
\textbf{Problem \tproblem.}

\smallskip

\noindent
\underline{Given:} A bipartite compatibility graph~$G(\agset,\rset,E)$, an 
agent~$\sag\in\agset$ seeking advice, its set of restrictions~$\restset$, the cost of removing
each restriction,
incompatibility set~$\Gamma$, and a budget~$\budget$.

\noindent
\underline{Requirement:} A set of restrictions~$\restset^*$
with~$\cost(\restset^*)\le \beta$ such that removal of  $\restset^*$ maximizes the gain in
probability~$\benefit(\restset^*)$.

\subsection{An Example}
\begin{figure}[htb]
\vspace*{-0.1in}
\centering
\includegraphics[width=.5\columnwidth]{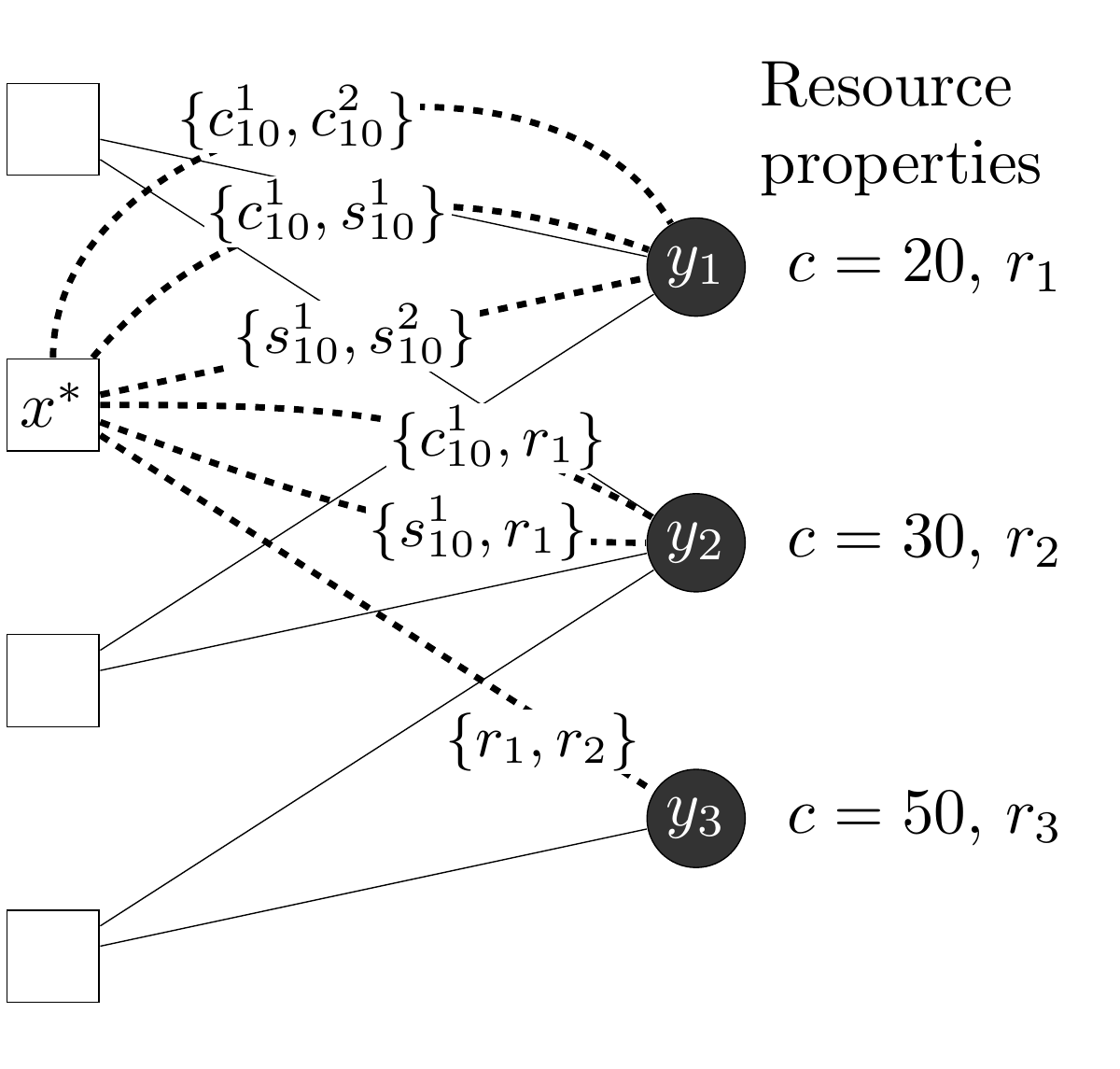}
\caption{A course-classroom example of matching advice framework. The
    agent~$\sag$ requires that the classroom capacity be at least~$40$ and
    located in region~$1$.
\label{fig:example}}
\vspace*{-0.35in}
\end{figure}

We use the following example of matching courses  to classrooms (see
Figure~\ref{fig:example}). Each classroom is a resource and each course (or
instructor) is an agent. Each classroom has two attributes: capacity and
region where it is located. Each course has restrictions such as the
required minimum capacity and desired regions.  Agent~$\sag$ prefers a
classroom of size at least~$40$ and regions in the order~$r_1>r_2>r_3$. In
the example of Figure~\ref{fig:example},~$\sag$ is incompatible with all
resources to begin with. To model the capacity restrictions, we discretize
the relaxation: we will assume that~$\sag$ relaxes the capacity constraint
in steps of~10. Accordingly, we have labels~$c^i_{10}$, where~$c$ denotes
the capacity and~$i$ denotes the step. For example, the capacity labels
associated with edge~$\{\sag,y_1\}$ are~$c^1_{10}$ and~$c^2_{10}$ as~$\sag$
must relax its capacity constraint by~$20$ for it to be compatible
with~$y_1$ with respect to capacity. There is an option to increase the
capacity by adding more seats (for a fee). Again, we assume that the
seating capacity can be increased in steps of~$10$. This is represented by
labels~$s^i_{10}$.  Relaxing capacity constraint by~$10$ is same as
increasing seating capacity by~$10$. Hence, as seen in
Figure~\ref{fig:example}, there are three ways for~$y_1$ to become
compatible with~$\sag$: reduce capacity requirement by~$20$ (remove
labels~$c^1_{10}$ and~$c^2_{10}$), increase seating capacity by~$20$
(remove labels~$s^1_{10}$ and~$s^2_{10}$), or reduce capacity requirement
by~$10$ and increase seating capacity by~$10$ (remove labels~$c^1_{10}$
and~$s^1_{10}$). For the region constraint, we have one label~$r_i$ for
every region~$i$. For~$y_3$ to be compatible with~$\sag$, both~$r_1$
and~$r_2$ must be removed. This is equivalent to saying that the
restriction that the classroom be located in regions $1$ or~$2$ is relaxed.

\subsection{Incompatibility types} 
\label{sse:incompat_type}
In this work, we consider advice frameworks with different forms
of incompatibility relationships.

\paragraph{\scmr{} incompatibility.} In an incompatibility set with this property, there
is exactly one choice for relaxing restrictions for each incompatible
resource. This means that in the incompatibility set~$\Gamma$, for every
incompatible resource~$y\in\sagrset$, there exists exactly one
resource--restrictions pair $(y,\restset')$. Note however that $|\restset'|$ may be $\geq 1$; 
i.e., more than one restriction may
need to be removed to make $y$ compatible with the agent.

\paragraph{\mcsr{} incompatibility.} In an incompatibility set with this
property, for each resource--restriction pair $(y,\restset')\in \Gamma$,
$|\restset'|=1$. This means that only one restriction needs to be removed
in order to make any resource compatible. However, it is possible that
there are multiple choices of restrictions to remove.

\smallskip
We also consider~\textbf{$\scsr$} incompatibility where for a resource, there is exactly one choice of one restriction to be removed to make it compatible. 
Similarly, we have~\textbf{$\mcmr$} incompatibility, a special case of which is the threshold-like incompatibility described below. 

\paragraph{Threshold-like incompatibility.}
This type of incompatibility is motivated by capacity and region
restrictions (as in Example~\ref{fig:example}).
In this case, the restrictions set~$\restset$ can be partitioned into~$\alpha$
blocks or \emph{attributes}~$\restset_\ell$,~$\ell=1,2,\ldots,\alpha$. In each
$\restset_\ell=\{r_{\ell,1},r_{\ell,2},\ldots,r_{\ell,t(\ell)}\}$, the restrictions can be
ordered~$r_{\ell,1}<r_{\ell,2}<\ldots<r_{\ell,t(\ell)}$,
where~$t(\ell)=|\restset_\ell|$. The incompatibility set~$\incset$ satisfies
the following property: $\forall (y,\restset')\in\incset$,
if~$r_{\ell,s}\in\restset'$, then, it implies
that~$r_{\ell,s+1}\in\restset'$ (if~$r_{\ell,s+1}$ exists).
In other words, if a restrictions
set $R'$ includes $r_{l,s}$, then it also includes
all higher elements $r_{l,s+1},r_{l,s+2}$, ... (provided they
exist).
Let~$r_{\ell,s}$ be the minimum 
element in~$\restset_\ell \cap \restset'$. It can be
considered as the \emph{threshold} corresponding to the~$\ell$th attribute
induced by the agent's restrictions. If a resource is incompatible with
regard to the~$\ell$th attribute, it means that the value of the resource
with respect to that attribute is less than $r_{l,s}$.
In the above example, the threshold for capacity is~$40$. Any classroom with
capacity less than~$30$ is below the threshold and hence is incompatible.

We also use abbreviated forms when necessary. For example, the short form
for $\scmr$ is $\scmrs$.


\section{Preliminaries}
\label{sec:preliminaries}
Here, we present some preliminary results regarding
maximum matching size and matching probability computation.
\begin{lemma}\label{lem:matching_size}
Let~$G$ denote the original compatibility graph and~$G'\ne G$
denote the compatibility graph obtained after some restrictions of
agent~$\sag$ are removed.
\begin{enumerate}[leftmargin=*,noitemsep,topsep=0pt]

\item Any maximum matching in $G'$ that is not a maximum matching in $G$
matches agent~$\sag$. 
In addition, the edge from~$\sag$ to the matched
resource is not in $G$.

\item The size of a maximum matching in $G'$
is at most one more than that of $G$.
\end{enumerate}
\end{lemma}

\iftoggle{arxiv}{
\noindent
\textbf{Proof (Idea):}
We use the simple fact that each 
new edge added to~$G'$ is incident on~$\sag$.
For details, see Section~\ref{sup:sec:preliminaries}
of the appendix.\hfill$\Box$
}{
\noindent
\textbf{Proof (Idea):}
We use the simple fact that each 
new edge added to~$G'$ is incident on~$\sag$.
For details, see \cite{TA+2022}.
\hfill$\Box$
}

\begin{definition}\label{def:matching_scenarios}
\textbf{Scenarios:}
Let $G$ and $G'$ denote respectively the original compatibility graph and the one that results after
some restrictions of agent~$\sag$ are removed.
There are two possible scenarios depending on
the sizes of maximum matchings of $G$ and $G'$.  
\begin{enumerate}[leftmargin=*,noitemsep,topsep=0pt]
\item {\bf Scenario 1.} Maximum matching size in
$G'$ is one more than that of $G$.
In this case,~$\sag$ is matched in all 
maximum matchings in $G'$.
Thus, in this scenario,
the probability that $\sag$ is
matched has the maximum possible value of 1.

\item {\bf Scenario 2.} Maximum matching size in $G'$ is
the same as that of $G$. In this
case, all maximum matchings of $G'$ which are not
maximum matchings in $G$ will have~$\sag$ matched to a resource.
\end{enumerate}
\end{definition}



\section{Hardness results}
\label{sec:hardness}
In this section, we present computational intractability results for
\tproblem.  To do this, we first define the decision version of \tproblem{},
which we denote by \tdproblem, as follows.

\medskip
\noindent
\textbf{Decision Version of \tproblem}~ (\tdproblem)$\:$:~

\smallskip
\noindent
\underline{Given:} A compatibility graph~$G(\agset,\rset,E)$, a special
agent~$\sag\in\agset$ seeking advice, its set of restrictions~$\restset$, the cost of removing
each restriction,
the incompatibility set~$\Gamma$, a budget~$\budget$, and a required
benefit~$\xbenefit$.

\noindent
\underline{Question:} Is there a set of restrictions~$\restset^*$
with~$\cost(\restset^*)\le \beta$ such that the gain in
probability~$\benefit(\restset^*)$ is at least~$\xbenefit$?

\smallskip
The following result establishes the complexity of \tdproblem{}
for the\\ \mcsrs{} advice framework.

\begin{theorem}\label{thm:or_hardness}
    \tdproblem{} is \cnp-hard for the
    \mcsrs{} advice\newline framework.
\end{theorem}

\iftoggle{arxiv}{
\noindent
\textbf{Proof (Idea):}~ 
Our reduction is from the \maxcover{} problem
which is known to be \cnp-complete \cite{GareyJohnson79}.
For details, see 
Section~\ref{sup:sec:hardness} of the appendix.
\hfill$\Box$
}{
\noindent
\textbf{Proof (Idea):}~ 
Our reduction is from the \maxcover{} problem
which is known to be \cnp-complete \cite{GareyJohnson79}.
For details, see \cite{TA+2022}.
\hfill$\Box$
}

\smallskip

Since the \mcsrs{} incompatibility is a special case
of threshold-like incompatibility, the following holds.
\begin{corollary}\label{cor:threshold_hardness}
    \tdproblem{} is \cnp-hard for the threshold-like advice framework.
\end{corollary}

\newcommand{\extor}{f_{\text{OR}}}
\newcommand{\extand}{f_{\text{AND}}}
\newcommand{\miset}[1]{M_i(#1)}
\newcommand{\misize}[1]{\omega_i(#1)}
\newcommand{\partitionset}{\Pi_\beta^\alpha}
\newcommand{\partition}{\mathbf{p}}

\section{Algorithms for Advice Frameworks}
\label{sec:algo}

\subsection{Notation}
\label{sec:alg_notation}

Let~$G$ denote the compatibility graph before any restriction of the
special agent $\sag$ is removed. For a subset of
restrictions~$A\subseteq\restset$, let~$G_A$ denote the compatibility graph
obtained by removing/relaxing~$A$ and let~$f(A)$ denote the number of new
maximum matchings in~$G_A$. By Part~(1) of Lemma~\ref{lem:matching_size},~$\sag$ is
matched in all these matchings. We call~$f(\cdot)$ the \emph{new matchings
count} function.  Using this notation,~$G$ corresponds to~$G_\varnothing$
and~$f(\varnothing)$ equals to $0$, where~$\varnothing$ is the empty set.
Note that the probability that~$\sag$ is matched in~$G$,~$p(G)$
(defined in Section~\ref{sec:definitions}) increases with~$f(\cdot)$.
We will use the standard definitions of monotone, submodular and
supermodular functions \cite{bach2013learning}. 
\iftoggle{arxiv}{
(For the reader's
convenience, these definitions are included in 
Section~\ref{sup:sec:alg_notation}
of the appendix.)
}{}
For simplicity, we use ``monotone'' to mean ``monotone non-decreasing''.

\subsection{Scenario Identification}
\label{sec:dm_scenario_id}
We recall from Lemma~\ref{lem:matching_size} and
Definition~\ref{def:matching_scenarios} that two scenarios are possible
when edges incident with~$\sag$ (meeting budget constraint) are added to
the compatibility graph. 
Further, in the case of Scenario~1, the
probability of matching~$\sag$ is~$1$; therefore, the probability
that~$\sag$ is matched needs to be estimated only for Scenario~2. In this
section, we will show an efficient method to (i)~determine whether
Scenario~1 exists, and if so, (ii)~find the set of restrictions 
to relax.
If the situation corresponds to Scenario~2, the algorithm returns the new incompatibility set $\Gamma'$ of $\sag$.

Our method crucially uses the Dulmage-Mendelsohn (DM) decomposition of the
node set of~$G$~\cite{dulmage1958coverings,pulleyblank1995matchings}.
Under this decomposition, any
maximum matching $M$ in a bipartite graph $G(\agset,\rset,E)$ defines a
partition of $\agset\cup\rset$ into three sets: odd ($\dmo$), even ($\dme$)
and unreachable ($\dmu$). 
A node $u\in\dme$ (respectively, $\dmo$) if there
is an even (odd) length 
\emph{alternating path}\footnote{Given a matching $M$, an alternating path between two nodes is a path in which edges in $M$ and edges not in $M$ alternate \cite{pulleyblank1995matchings}.
The length of such a path is the number of edges in the path.}
in $G$ from an unmatched node to
$u$.  A node $u\in\dmu$, that is, it is unreachable, if there is no
alternating path in $G$ from an unmatched node to $u$. 
We will use the following well-known results.
\begin{lemma}[Irving et al.~\cite{irving2006rank}]\label{lem:dm}
Consider a bipartite graph $G(\agset,\rset,E)$ and let $\dme$, 
$\dmo$ and $\dmu$ be defined as above with respect to a maximum
matching $M$ of $G$.
\begin{enumerate}
\item The sets $\dme$, $\dmo$ and $\dmu$ form a partition of 
$\agset\,\cup\,\rset$, and this partition is 
\underline{independent} of the maximum matching.
\item In any maximum matching $M$ of $G$ the following hold.
\begin{enumerate}
  \item $M$ contains only $\dmu\dmu$ and $\dmo\dme$ edges.
  \item Every vertex in $\dmo$ and every vertex in $\dmu$ is matched by $M$.
  \item $|M|$ = $|\dmo|+|\dmu|/2$.
\end{enumerate}
\item There is no $\dme\dmu$ edge or $\dme\dme$ edge in $G$.
\end{enumerate}
\end{lemma}

\begin{lemma}\label{lem:sc1}
Let~$M$ be a maximum matching and~$x\in \agset\,\cap\,\dme$. Adding
edge~$\{x,y\}$, where $y\in\rset$ is an incompatible resource,
increases the matching size iff ~$y\in \dme$.
\end{lemma}
\iftoggle{arxiv}{
A proof of this lemma appears in Section~\ref{sup:sec:dm_scenario_id} of
the appendix. 
}{
A proof of this lemma appears in \cite{TA+2022}.  
}
The method to identify Scenario~1 is described in
Algorithm~\ref{alg:sce_one}.

\begin{algorithm}[ht]
\caption{\small{Detecting Matching Scenario~1 and updating the incompatibility set}\label{alg:sce_one}}
\DontPrintSemicolon
\BlankLine
\SetKwInOut{Input}{Input}
\SetKwInOut{Output}{Output}
\Input{
 Agents~$\agset$,
Resources~$\rset$,
compatibility graph~$G$, special agent~$\sag$,
its incompatibility set~$\incset$ and budget~$\budget$.
}
\Output{Decide if Matching Scenario~1 has occurred or not. If not, output the
new incompatibility set~$\incset'$ that accounts for the budget.}
Set $\Gamma' = \emptyset$\;
Compute the DM-decomposition of $\agset\cup\rset$ into $\mathcal{O}$, $\mathcal{U}$, and $\mathcal{E}$\;
\For{each $(y,\restset')\in\incset$}{ \label{line:incompatibility}
    \If{$\cost(\restset')\le\budget$}{
        \If{$y\in\mathcal{E}$}
        {\Return ``Matching Scenario~1 detected'' and $R'$\;}
        \Else{
        $\incset'\leftarrow\incset'\cup\{(y,\restset')\}$.}
    }
}
\Return ``Matching Scenario~2 detected" and $\incset'$
\end{algorithm}
\paragraph{Correctness of Algorithm~\ref{alg:sce_one}.} We will now show
that the algorithm detects Matching Scenario~1, if it exists. 
We note that
this scenario can happen if and only if the following two conditions are
met: (i)~there exists a resource~$y\in\dme$ and
(ii)~there exists a resource-restrictions pair~$(y,R')$ such
that~$\cost(R')\le\budget$. The first condition is due to
Lemma~\ref{lem:sc1}, while the second follows from the budget constraint.
The algorithm checks for precisely these conditions. Hence, it
detects Scenario~1 if it exists. 
Also, note that the algorithm filters out
resource-restrictions pairs that do not meet the budget constraint.


\begin{lemma}
Algorithm~\ref{alg:sce_one} runs in time $O(m\sqrt{n}+ |\incset|))$,
where $n$ and $m$ are the number of nodes and
edges in $G$ and $\Gamma$ is the incompatibility set of $\sag$.
\end{lemma}

\noindent
\textbf{Proof:}~
To compute the DM-decomposition, we need to first compute a maximum
matching $M$. This takes $O(m\sqrt{n})$ time using the Hopcroft-Karp algorithm \cite{CLRS-2009}.
Given $M$, computing the DM-decomposition can be done in~$O(m)$
time~\cite{pulleyblank1995matchings,irving2006rank}. 
Using this decomposition, checking whether a node $y$ is
in $\dme$ can be done in $O(1)$ time.
Since for each $(y,R') \in \Gamma$, the value $\rho(R')$
can be precomputed, checking whether $\rho(R') \leq \beta$
can also be done in $O(1)$ time.
Thus, each iteration of the for loop in
Line~\ref{line:incompatibility} uses $O(1)$ time.
Hence, the total time used by the loop is~$O(|\incset|)$.
Therefore, the running time of the algorithm is 
$O(m\sqrt{n}+ |\incset|))$. \QED

\begin{algorithm}[t]
\caption{{\small Greedy algorithm for \mcsrs{} 
corresponding to Matching Scenario~2 with uniform probability of choosing a maximum
matching and uniform cost for relaxing restrictions} \label{alg:or_greedy}}
\DontPrintSemicolon
\BlankLine
\SetKwInOut{Input}{Input}\SetKwInOut{Output}{Output}
\Input{Agents~$\agset$,
Resources~$\rset$, 
compatibility graph $G$, special agent $\sag$,
its incompatibility set $\incset$, budget $\budget$ and an oracle
for the probability $p(G)$ that~$\sag$ is matched in the compatibility 
graph~$G$.
}
\Output{Set of restrictions~$A^*\subseteq \restset$,~$|A^*|\le\budget$}
$A^*=\varnothing$.\;
\While{$|A^*|<\beta$}{
    $r^*=\argmax_{r\in\restset}p(G_{A^*\cup\{r\}})$.\;
    $A^*\leftarrow A^*\cup\{r^*\}$
    and~$\restset\leftarrow\restset\setminus\{r^*\}$.\;
}
\Return $A^*$
\end{algorithm}


\subsection{\mcsr{}}
\label{sec:disjunction} 
Here, we consider the \mcsr{} incompatibility framework where
any resource can be made compatible with the removal of exactly one
restriction. We will assume throughout that the cost of removing any
restriction is~$1$ and that the maximum matching algorithm samples matchings
uniformly from the space of all maximum matchings. We note that for the latter
case, the probability of~$\sag$ being matched is the fraction of the maximum matchings of the given compatibility
graph in which $\sag$ is matched.
\begin{lemma}
\label{lem:or_submodular}
Consider the \mcsrs{}
incompatibility. Then, for Matching Scenario~2, the new matching count
function~$f(\cdot)$ is monotone submodular.
\end{lemma}
\iftoggle{arxiv}{
For a proof of the above lemma, see Section~\ref{sup:sec:disjunction}
of the appendix.
}{
For a proof of the above lemma, see \cite{TA+2022}.
}
Since $f$ is monotone submodular, we can use the greedy algorithm that
iteratively picks a restriction with the highest benefit-to-cost ratio
to relax~\cite{nemhauser1978analysis}. Since each addition has
the same cost (namely,~1), the highest benefit-to-cost ratio is achieved by
a restriction that has the highest benefit.  The resulting algorithm, which provides an approximation for the
\mcsrs{} case, is shown
as Algorithm~\ref{alg:or_greedy}. Note again that in the algorithm, 
we are using the fact that~$p(\cdot)$ increases with $f(\cdot)$.
The following result is again due to the fact that~$f$ is a monotone
submodular function;
\iftoggle{arxiv}{
see Section~\ref{sup:sec:disjunction} of the 
appendix for a proof of the following result.
}{
see \cite{TA+2022}
for a proof of the following result.
}

\newcommand{\nonmatch}{\omega^-_{\text{old}}}
\newcommand{\oldmatch}{\omega^+_{\text{old}}}
\newcommand{\Gopt}{G_{\text{opt}}}
\newcommand{\matchcurr}{\omega^+_{G'}}
\newcommand{\matchopt}{\omega^+_{\Gopt}}
\newcommand{\pcurr}{p(G')}
\newcommand{\popt}{p(\Gopt)}

\begin{theorem}\label{thm:or_greedy}
Consider the \mcsrs{} incompatibility.
Suppose each restriction has the same removal cost  and the maximum
matchings of the compatibility graph~$G$ are chosen from the uniform
distribution. Then, given an oracle for computing the probability 
$p(\cdot)$, Algorithm~\ref{alg:or_greedy} provides a
solution to the \tproblem{} problem with cost at most~$\beta$ and benefit at 
least~$(1-1/e)$ of the optimal solution.
\end{theorem}

\smallskip
\noindent
Suppose the incompatibility set satisfies single restriction and
single choice properties. Then, it can be shown that~$f$ is monotone and modular,
in which case, the greedy algorithm is optimal
\cite{edmonds1971matroids}.
\begin{corollary}\label{cor:or_and_greedy}
Consider the \tproblem{} problem under single restriction and single choice
incompatibility. Suppose each restriction has the same removal cost
and the maximum matchings of the compatibility graph~$G$ are chosen from
the uniform distribution. Then, Algorithm~\ref{alg:or_greedy} is optimal.
\end{corollary}

\iftoggle{arxiv}{
\noindent
\textbf{Proof.}~
The proof is similar to that of
Theorem~\ref{thm:or_greedy}. In the proof of Lemma~\ref{lem:or_submodular},
we note that~$|B_y|=1$ since the incompatibility set satisfies the single
choice property. Therefore, the set~$C_4$ is empty or~$\omega_4=0$.
Hence,~$f(B)+f(B')=f(B\cup B')+f(B\cap B')$, and therefore,~$f$ is modular.
Hence, the greedy strategy in Algorithm~\ref{alg:or_greedy} gives an optimal
solution~\cite{edmonds1971matroids}. \QED
}{
\noindent
\textbf{Proof (Idea):}~
We show that the function $f$ is modular, that is,
it is both submodular and supermodular
(see \cite{TA+2022}).
Hence, the greedy strategy in Algorithm~\ref{alg:or_greedy} gives an optimal
solution~\cite{edmonds1971matroids}.
\hfill$\Box$
}

\subsection{Threshold-like Incompatibility}
\label{sec:threshold}
We now describe an algorithm for finding an optimal solution to the
\tproblem{} problem for threshold-like incompatibility. We assume that the budget $\beta$ and 
cost of removing each restriction are non-negative integers.
Let~$\restset=\biguplus_{1\le\ell\le\alpha}\restset_\ell$ be a partition of
variables where each part contains variables corresponding to values of an
attribute. We say that an $\alpha$-tuple
$(\beta_1,\beta_2,\ldots,\beta_\alpha)$ of non-negative integers is an
$\alpha$-partition of the budget~$\beta$ if $\sum_{\ell=1}^{\alpha}
\beta_{\ell} = \beta$.  Let~$\partitionset$ denote all the
$\alpha$-partitions of~$\beta$. Algorithm~\ref{alg:thresh} exhaustively
explores all possible budget allocations to the attributes. Once the budget
is allocated, the best solution among the restrictions in
each~$\restset_\ell$ can be computed using a binary search. We identify the
least restriction~$r^*_\ell$ in~$\restset_\ell$ such that the sum of costs
of all~$r\ge r^*_\ell$ in~$\restset_\ell$ are removed. Unlike the previous
cases, this algorithm does not assume uniform cost or uniform probability
of picking a maximum matching.  

\begin{algorithm}[!htb]
\caption{{\small Algorithm for threshold-like incompatibility corresponding to
Matching Scenario~2} \label{alg:thresh}}
\DontPrintSemicolon
\BlankLine
\SetKwInOut{Input}{Input}
\SetKwInOut{Output}{Output}
\Input{
Resources~$\rset$, agents~$\agset$,
compatibility graph~$G$, special agent~$\sag$,
its incompatibility set~$\incset$, budget~$\budget$ 
and a probability oracle~$p(\cdot)$.
}
\Output{Set of restrictions~$A\subseteq \restset$ with $\cost(A)\le\budget$}

$p^*=p(G)$ and $A^*=\varnothing$\;
\For{each $(\beta_1,\beta_2,\ldots,\beta_\alpha)\in\partitionset$}{
    \For{$\ell=1,2,\ldots,\alpha$}{
        $r^*_\ell =\argmin_{r\in\restset_\ell}\sum_{r'\ge r}\cost(r')\le
        \budget_\ell$. \label{alg:threshold:min}\;
        Let~$A_\ell=\{r\mid r\in\restset_\ell,\,r\ge r^*_\ell\}$.\;
    }
    Let~$A=\bigcup_\ell A_\ell$.\;
    \lIf{$p(G_A)>p^*$}{$A^*=A$, $p^*=p(G_A)$}
}
\Return $A^*$
\end{algorithm}

\vspace*{-0.2in}

\begin{theorem}\label{thm:threshold}
For \tproblem{} with threshold-like incompatibility
where the budget $\beta$ and the cost of removing each restriction are non-negative integers,
given an oracle for probability~$p(\cdot)$, Algorithm~\ref{alg:thresh} provides an optimal solution in  $O(\beta^\alpha\log|\restset|)$ calls to the probability~$p(\cdot)$ computing oracle, where
$\restset$ is the restrictions set of special agent~$\sag$, and~$\alpha$ is the number of blocks in~$\rset$.
\end{theorem}

\iftoggle{arxiv}{
Our proof of the above result appears in Section~\ref{sup:sec:threshold} of the appendix.
}{
Our proof of the above result appears in \cite{TA+2022}.
}

\newcommand{\coll}{\mathcal{R}_A}

\newcommand{\rcoll}{\mathcal{R}}
\vspace*{-0.1in}
\section{Computing Matching Probability}
\label{sec:prob_compute}

A crucial component of the
advice framework is to estimate the probability that a maximum matching
chosen uniformly randomly from the set of all maximum matchings
includes~$\sag$. This problem is closely related to a computationally
intractable (technically, \cnump-hard) problem, namely counting the number
of maximum matchings in bipartite graphs
\cite{valiant1979complexity,Jerrum-1987}. In our case, this probability
computation must be repeatedly performed each time a possible solution is
to be evaluated. Our goal here is to reduce the number of such
computations.  We will show that under certain independent sampling of
matchings, one can precompute a relatively small number of probabilities
that can be used to find the probability of~$\sag$ being matched after
relaxing any set of restrictions.

Suppose the set \rset{} of resources can be partitioned into~$\eta$
blocks~$\rset= \rset_1\uplus\rset_2\uplus\cdots\uplus\rset_{\eta}$ such
that for any set of restrictions~$A\subseteq\restset$ and any
block~$\rset_\ell$, relaxing~$A$ either makes all resources in~$\rset_\ell$
compatible or none of its resources compatible with~$\sag$.
Let $G_A$ denote the compatibility graph after the restrictions in $A$ are removed.
Under the
assumption that the matchings are sampled 
independently of one another from $G_A$, to
compute the probability after relaxing~$A$, it is enough to know the
probability value $p_{\ell}$ that~$\sag$ is matched to a resource in~$\rset_\ell$, $1\le\ell\le\eta$ when sampled from all possible maximum
matchings. Let~$p_0$ denote the probability that 
$\sag$ is not matched.
Let~$\rcoll(A)$ denote the set of blocks whose resources become compatible
with~$\sag$ after relaxing~$A$. Then, probability that~$\sag$ appears in
a maximum matching after relaxing~$A$ is given by
    $\frac{{\sum}_{\srset_\ell\in\rcoll(A)} {p_\ell}}
{p_0+{\sum}_{\srset_\ell\in\rcoll(A)} {p_\ell}}$.
The justification for the summation used here
is that every maximum matching
containing~$\sag$ has exactly one resource matched to it. Therefore, the
events that~$\sag$ is matched to a resource
in~$\rset_\ell$,~$1\le\ell\le\eta$ are disjoint.

We note that the number of resources~$m$ is a trivial upper bound
for~$\eta$, the number of blocks. 
In the case of threshold-like
incompatibility, another upper bound can be specified. For budget~$\beta$ and number
of attributes~$\alpha$, the number of optimal 
solutions is bounded
by~$\budget^\alpha$ (Algorithm~\ref{alg:thresh}). This serves as an upper bound for~$\eta$.

\begin{figure}[t]
\centering
\small
\includegraphics[width=0.272\columnwidth]{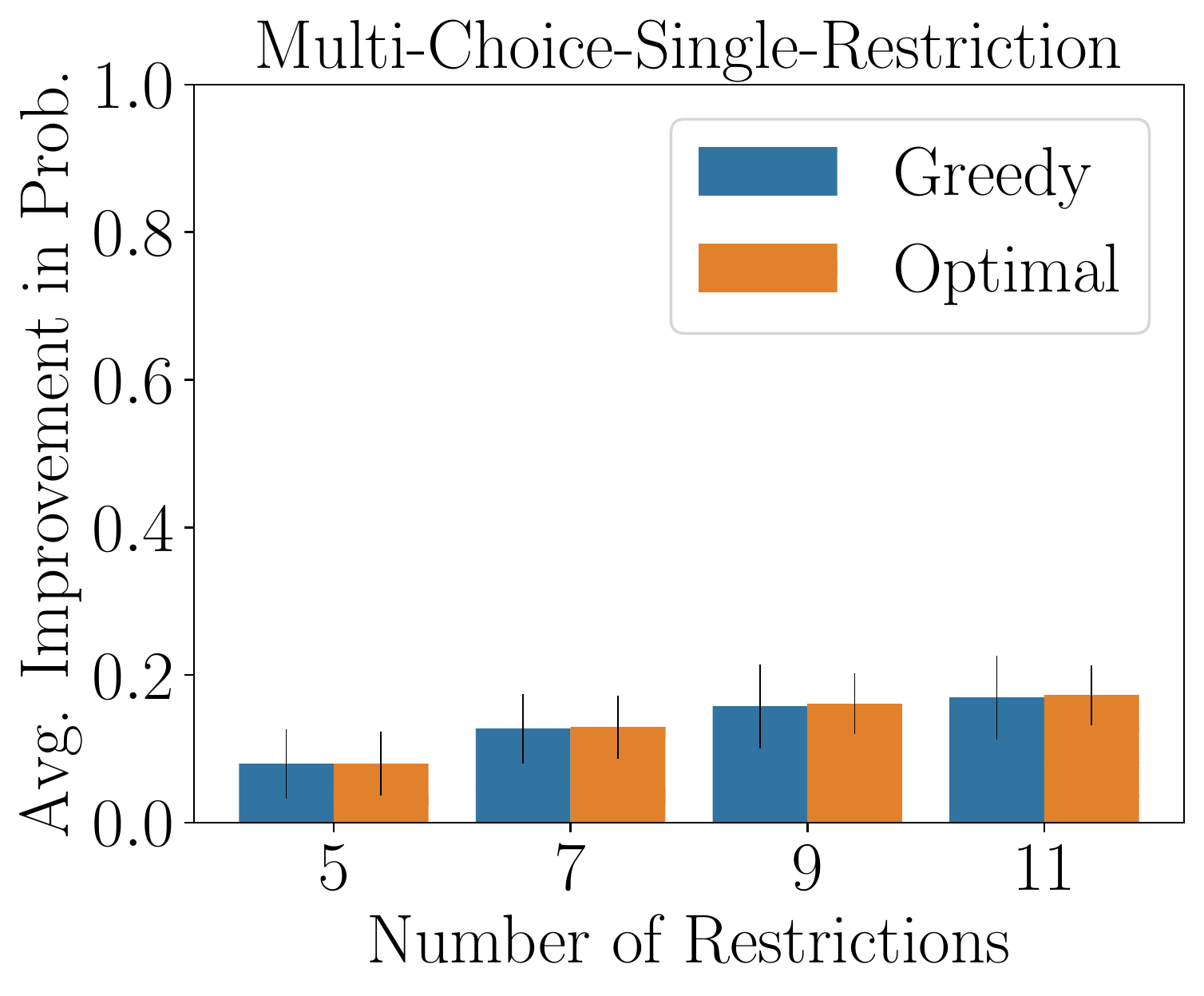}
\includegraphics[width=0.234\columnwidth]{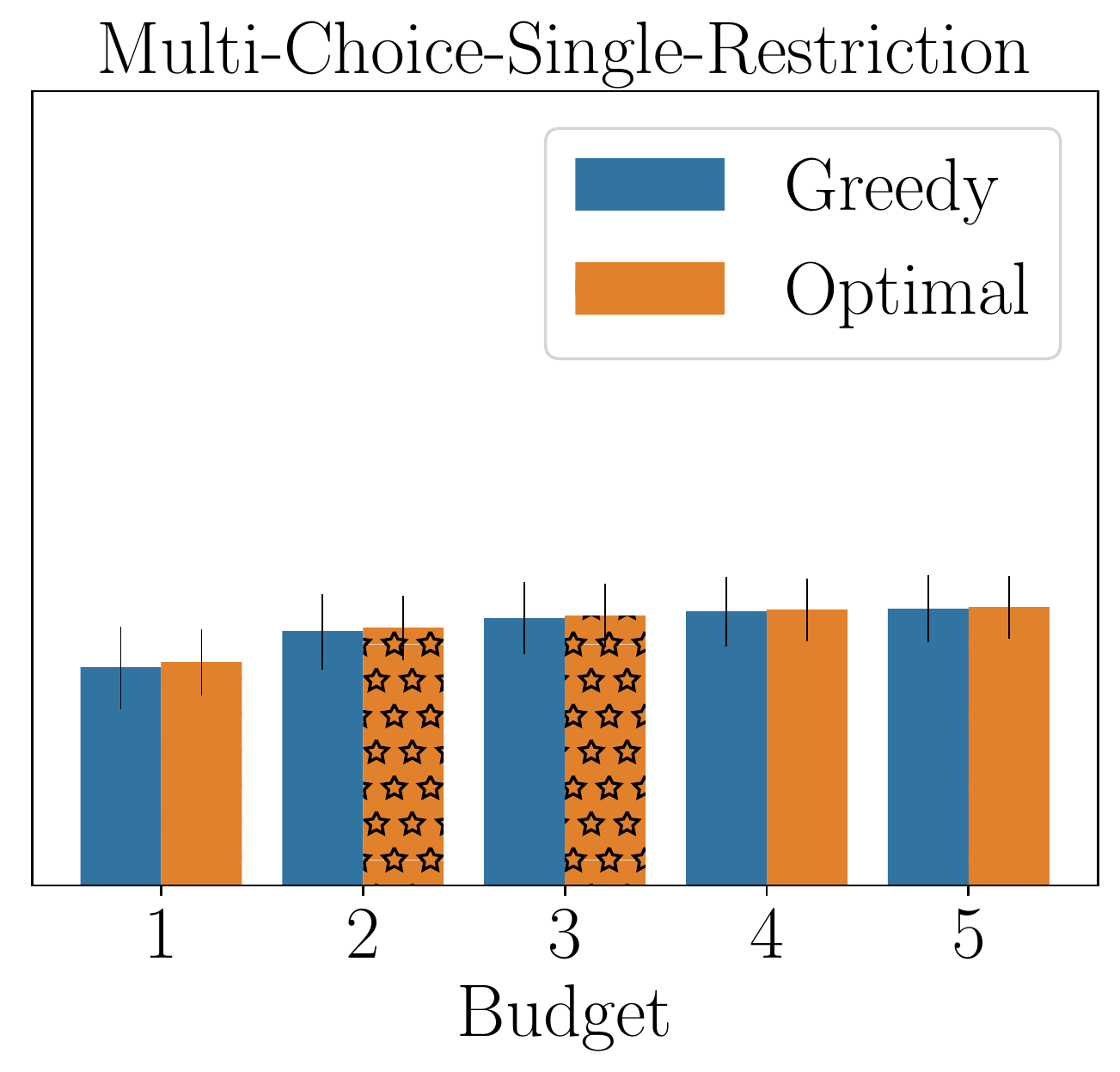}
\includegraphics[width=0.234\columnwidth]{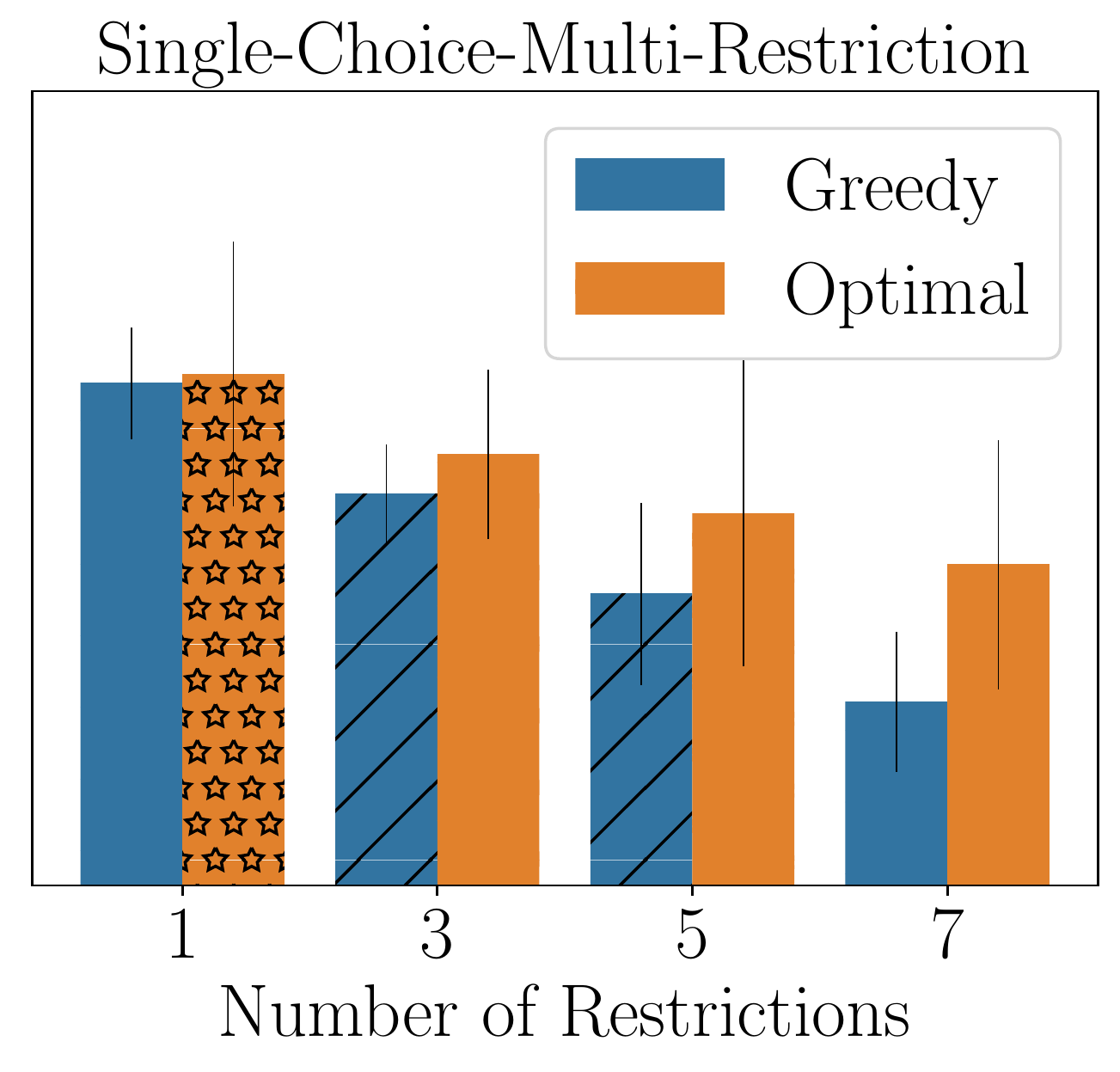}
\includegraphics[width=0.234\columnwidth]{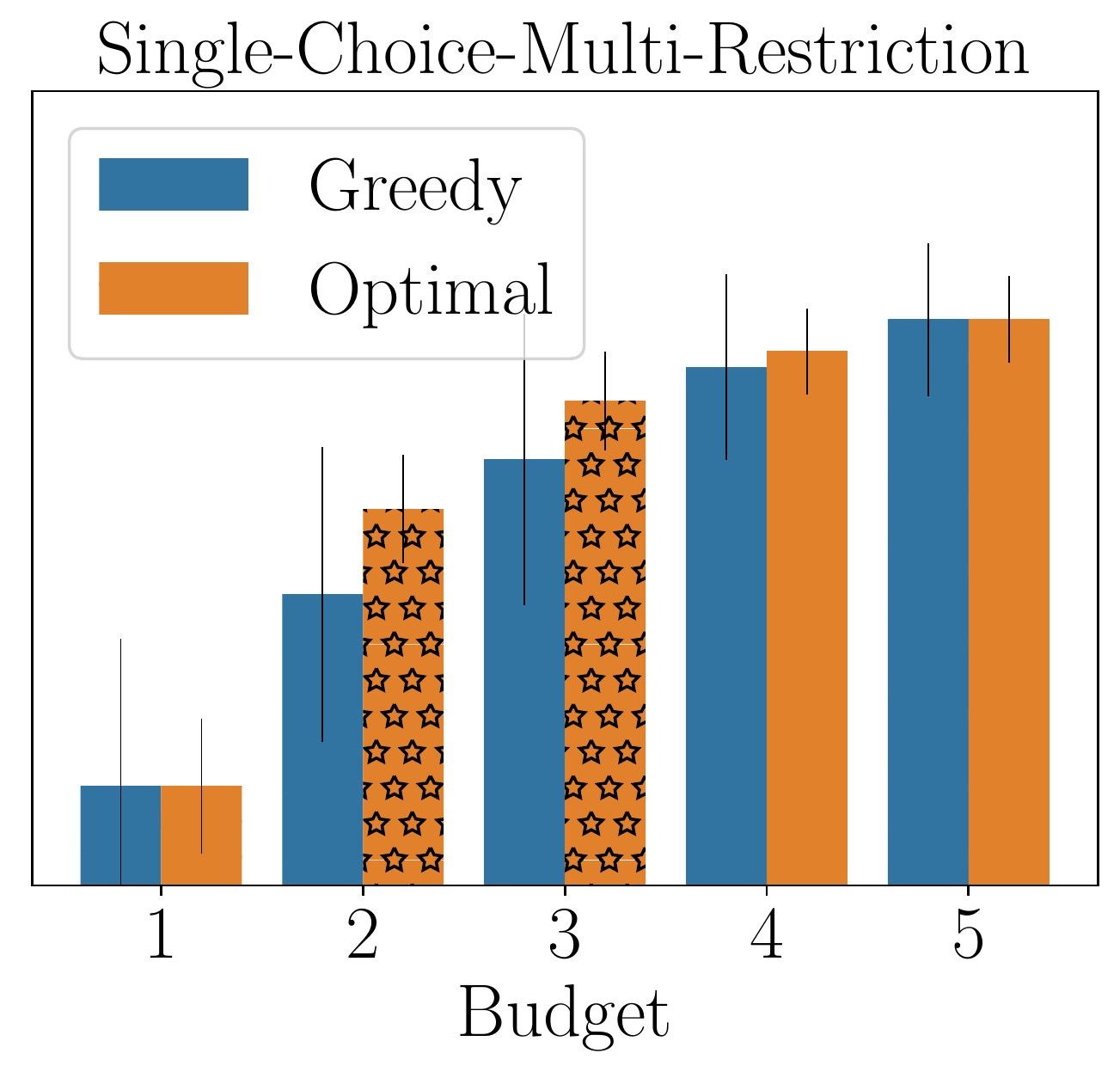}
\caption{The results for \mcsrs{} and \scmrs{}
on random bipartite graphs for varying number of
restrictions and budget. 
The range of values on the
y-axis of all plots are the same.}
\vspace*{-0.2in}
\label{fig:synth}
\end{figure}   

\section{Experimental Results}
\label{sec:experiments}
\newcommand{\costi}{Cost-\uppercase\expandafter{\romannumeral 1\relax}}
\newcommand{\costii}{Cost-\uppercase\expandafter{\romannumeral 2\relax}}

\newcommand{\pv}{\textsc{PassVac}}
\newcommand{\cc}{\textsc{CoCl}}
\newcommand{\ccz}{\textsc{CoCl-zc}}
We experimented extensively on real-world and synthetic datasets to
evaluate our algorithms for the advice frameworks considered.

\paragraph{Datasets.}
We considered a family of synthetic graphs and real-world datasets.  We
used Erd\"{o}s-Renyi random bipartite graphs \cite{EK-2010}~$G(n,p)$ for
experiments to evaluate the greedy algorithms for the \mcsrs{}
and \scmrs{} incompatibility frameworks.  For the threshold-like
incompatibility, we considered two real-world datasets. The first 
dataset is the \emph{Course-Classroom} (\cc) dataset. This comes from a university\footnote{Bar-Ilan University, Ramat Gan, Israel.} for the year~2018--2019\footnote{Dataset is available at \url{https://github.com/yohayt/RAR\_EUMAS2022}}.
In the experiments we focused on a two-hour slot on a specific day of the
week (Tuesday), and used all the courses that are scheduled in this time
slot and all available rooms.  There are~$144$ classrooms and~$154$
courses.  Each classroom has four attributes: its \emph{capacity}, the
\emph{region} to which it belongs, whether it allows students with
physical disability and whether it allows students with hearing disability.
Following the COVID-19 epidemic, additional features were added to the
classes such as whether the class has facility for
remote learning (\url{https://zoom.us/} in this case). If the classroom
has no feature for remote learning, then the
teacher must bring the required equipment. Another feature was
flexibility
to add chairs to a class to increase its capacity.  So, we have the
attribute-augmented dataset~\ccz{} with the following extra features
compared to \cc: (i)~adding chairs as an alternative to reducing
capacity, (ii)~remote learning in the classroom, and
(iii)~portable Zoom equipment as an alternative to~(ii). Note that \cc{}
corresponds to \scmrs{} threshold-like incompatibility, while \ccz{}
corresponds to \mcmrs{} threshold-like incompatibility. Even though 
assigning classrooms to courses is
well-studied~\cite{Phillips2015integer} we did
not find any publicly available dataset.  The
\emph{Children Summer Vacation Activities} or \emph{Passeport
Vacances}~(\pv)~\cite{varone2019dataset} corresponds to online registration
for assigning holiday activities to children. There are three attributes --
minimum and maximum permissible age for participation with ranges. In
addition, each child has restrictions as to which activity they would like
to participate in. The minimum and maximum age restrictions each correspond
to a threshold function. Note that the activity might be either too trivial
for the child if the minimum age is relaxed, or the child may not fully
understand the activity if the maximum age is relaxed.  The numbers of
children and activities are 634 and 533, respectively. We focused on one of
the vacations in the dataset. In this vacation, there were 249 activities.

\begin{figure*}[hbt]
\centering
\small
\includegraphics[width=.34\textwidth]{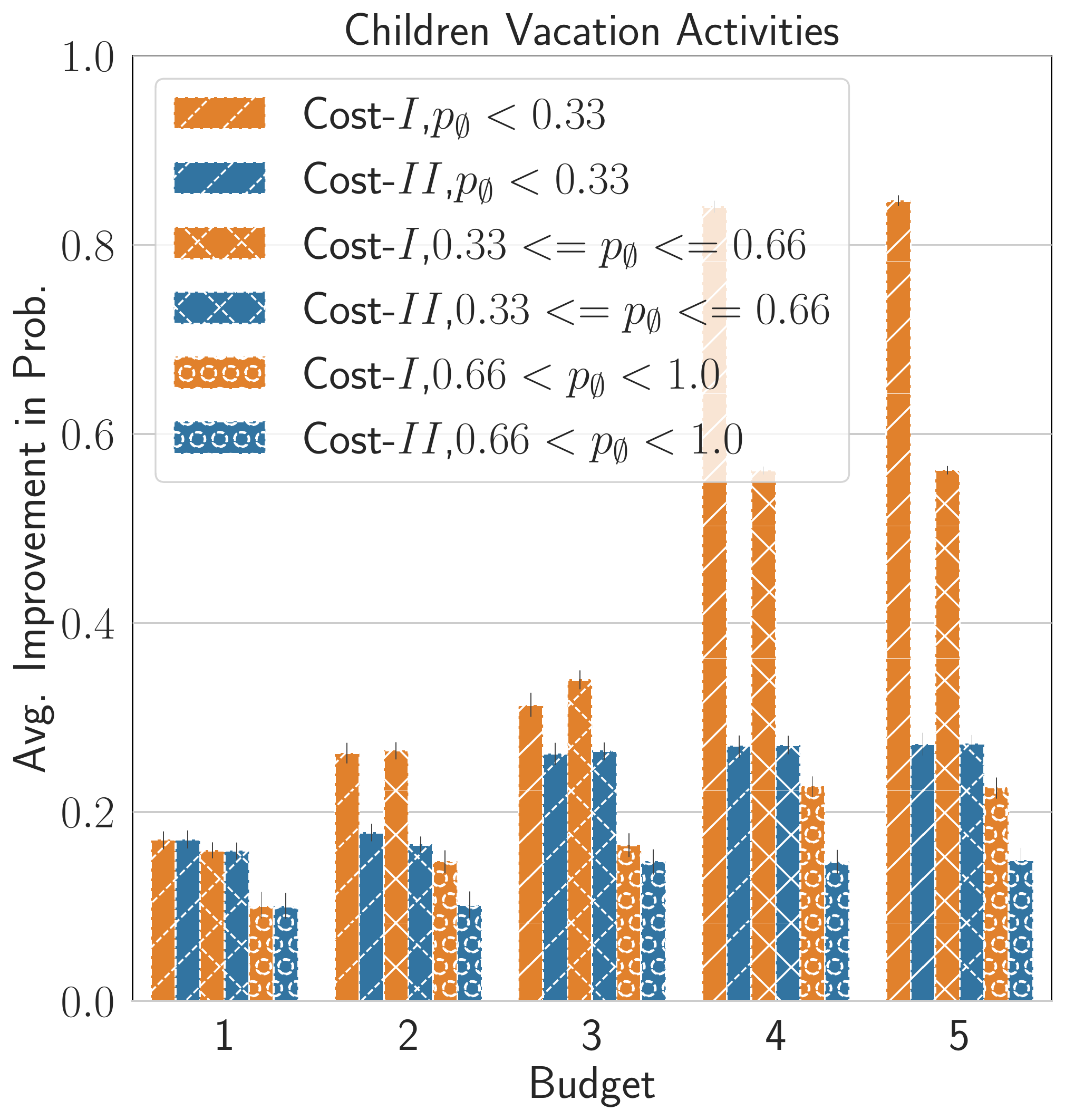}
\includegraphics[width=.3\textwidth]{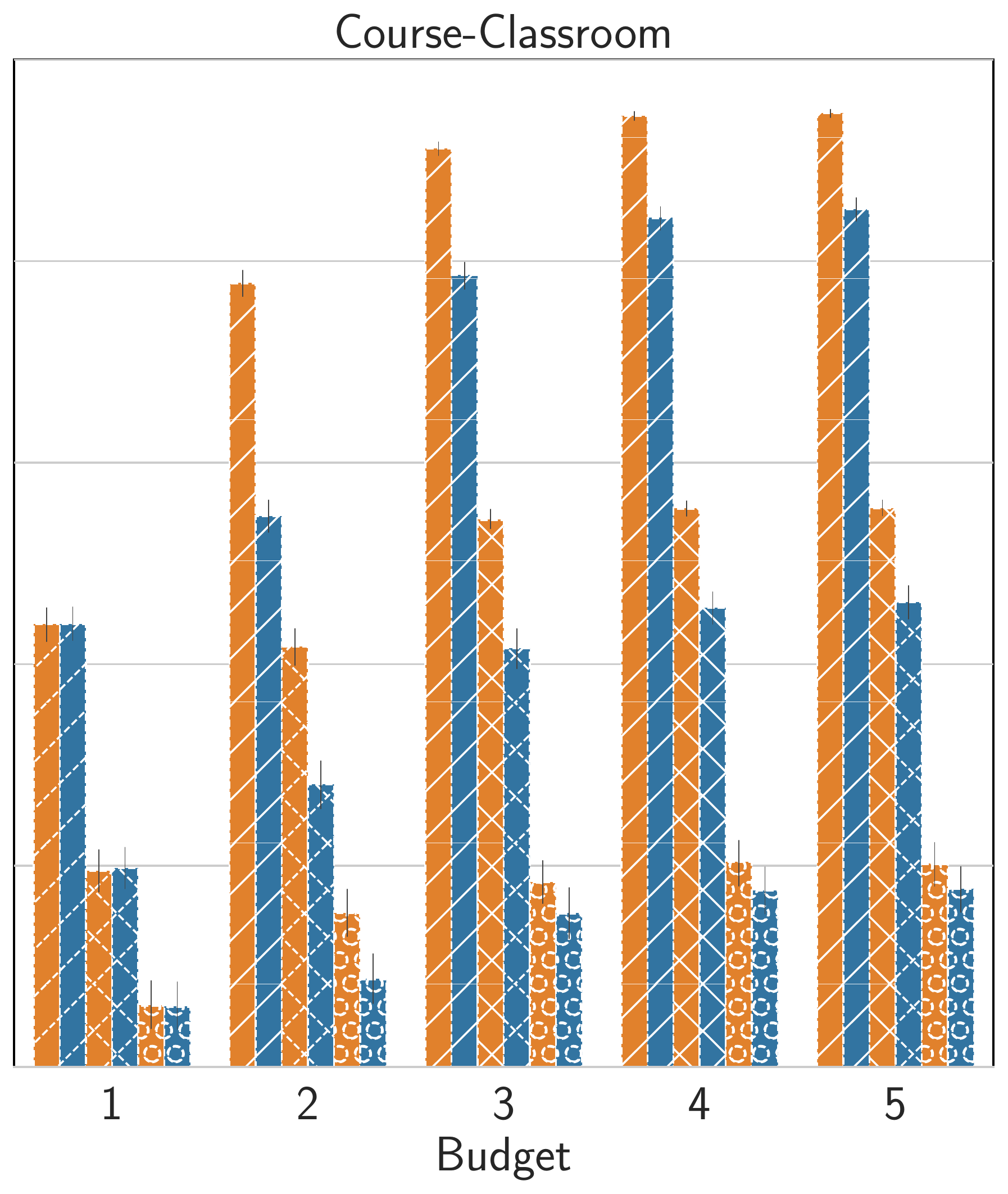}
\includegraphics[width=.298\textwidth]{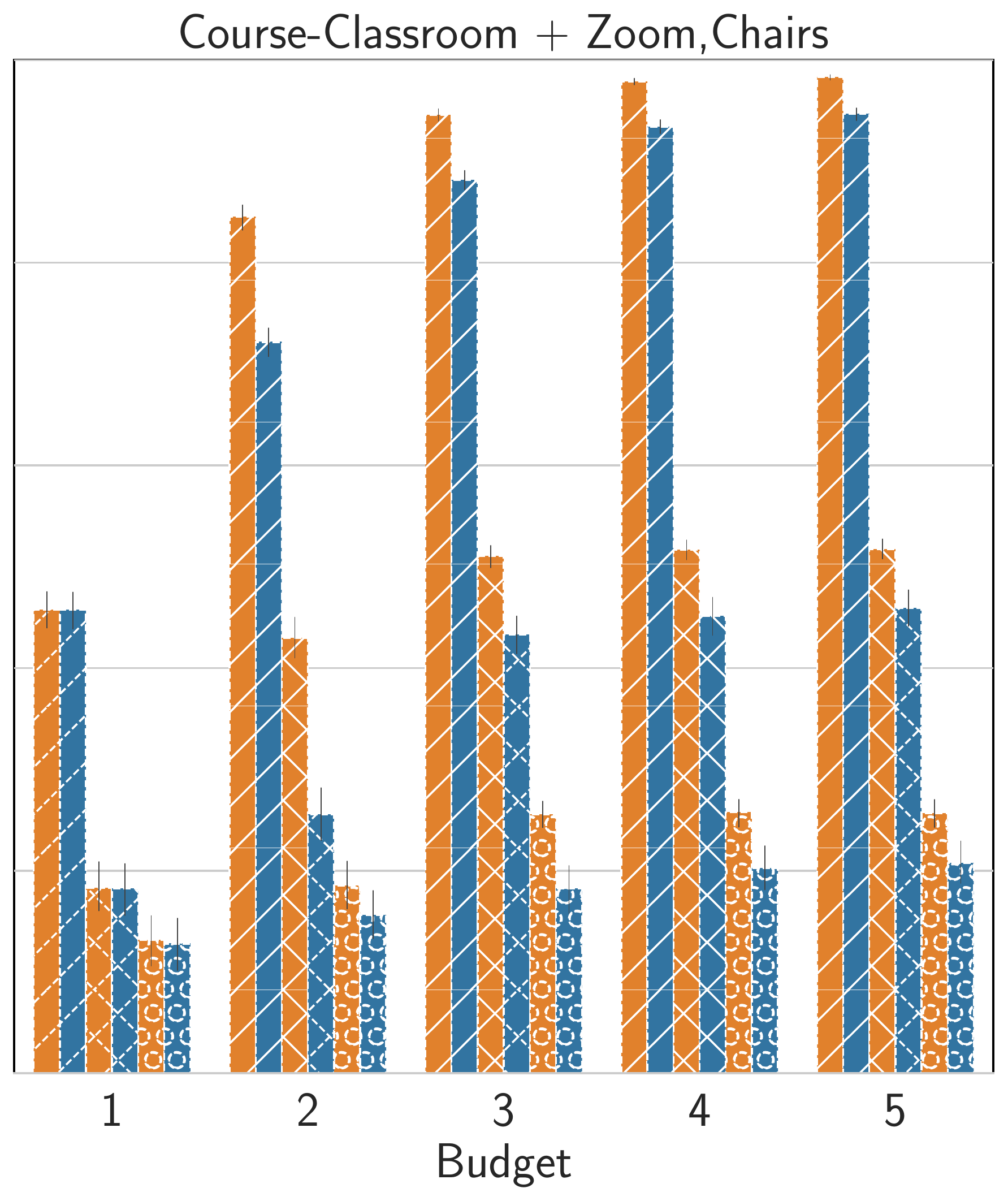}

\caption{The benefit obtained by removing restrictions for the
threshold-like incompatibility on (i)~\pv, (ii)~\cc{}, and (iii)~\ccz{}.
For analysis, we have partitioned the agents based on their original
estimated probability of matching~$p_\varnothing$.}
\vspace*{-0.2in}
\label{fig:real}
\end{figure*}

\paragraph{Probability computation.} 
In all the experiments, the probability that the special agent~$\sag$ is
matched was estimated in the following manner. A random maximum matching
was generated by first randomly permuting the set of agents and using the
resulting compatibility graph as input to the Hopcroft-Karp
algorithm~\cite{Hopcroft-Karp-1973}. Each time, 1000 such maximum
matchings were generated. The probability of~$\sag$ being matched is simply
the ratio of total number of matchings in which $\sag$ is matched to~1000.

\paragraph{\mcsrs{} with synthetic graphs.}
We generated 100 random bipartite graphs, each with 40 agents and 20
resources. Each edge has the probability of~0.2 to be matched. Then an
additional agent was generated as~$\sag$. A subset of restrictions was
generated randomly for each resource. 
If the generated set is empty,
then that resource would be made compatible with the agent. We experimented
extensively on synthetic graphs to evaluate the greedy approach of
Algorithm~\ref{alg:or_greedy} by varying the size of the restrictions
set and budget size.  We used
exhaustive search to obtain a pseudo-optimal solution (since the
probabilities are only estimates) and compared it with the greedy solution.
For each instance, we ran~100 experiments of finding sets of restrictions to be removed. 
The results are in the first two parts of Figure~\ref{fig:synth}. 
We varied the maximum number of restrictions allotted per
resource from~$2$ to~$4$. In the top left plot, we fixed the budget~$\budget$ to $2$ and varied 
the number of restrictions from~$5$ to~$11$. In the bottom left plot, we fixed the number of restrictions
to $19$ and varied the budget~$\budget$ from~$1$ to~$5$.
We observe that in each case, the greedy algorithm closely matched the performance
of the pseudo-optimal solution. We note the gain is high for small budgets
as only one restriction per resource needs to be removed to make it
compatible. Therefore, increasing the budget only increases the gain
marginally. Increasing the number of restrictions does not have much effect
on the benefit.

\paragraph{\scmrs{} incompatibility with synthetic graphs.}
Here,  we apply the greedy
algorithm (Algorithm~\ref{alg:or_greedy}). It is known to have
performance guarantee of~$\gamma$ times the best solution given the budget,
where~$\gamma$ is the submodularity ratio~\cite{bian2017guarantees}.
Again, we used exhaustive search to obtain a pseudo-optimal solution
and compared it with the greedy solution.  The experiment design is
similar to that of
the \mcsrs{} case.  In the third and fourth
parts of Figure~\ref{fig:synth}, the
results are presented for varying sizes of restrictions and budget. The
number of restrictions per resource is at most~$4$. We note that unlike
the \mcsrs{} case, the probability of being matched decreases with
increase in the number of restrictions; this is because all
the restrictions corresponding to a resource must be removed for it to
become compatible. Also, increasing the budget provides significant 
benefit in this case as many more restrictions must be removed for
resources to become compatible compared to the \mcsrs{} case.

\paragraph{Threshold-like incompatibility with real data sets.} 
For \cc{}
dataset, we used 140 courses out of the 154 available and for each agent,
each cost function, and  each budget value, we have~100 replicates. For \pv{}, we used 603 children out of the 634 children.
For each agent, 
each cost function, and  each budget value, we have~30
replicates. For both \pv{} and \cc{} datasets, we used two cost schemes:
\costi{} is the uniform cost function where all
attribute values have cost~$1$ and~\costii{} is a linear cost function,
where, for a given attribute, the cost of removing the first 
restriction is~$1$, the second is~$2$, and so on. Therefore, if~$t$
labels corresponding to an
attribute are removed, the cost incurred is~$t\,(t+1)/2$ (the
more the \prin{} deviates from the threshold set by the agent, the
higher the regret or penalty). Here, we considered multiple agents, one
at a time in our analysis. These agents were categorized based on their
initial probability of being matched, $p_\varnothing$: (i)~$[0,1/3)$,
(ii)~$[1/3,2/3)$, and (iii)~$[2/3,1)$. The results shown in
Figure~\ref{fig:real} are discussed below.

\noindent
\underline{Application specific observations.} For \cc, hearing disability
feature is typically the first to be relaxed. This seems to suggest that
for the number of students with hearing disability, the number of
classrooms which can accommodate their needs is not adequate. In the \pv{}
case, the abrupt increase in probability was due to a large number of
activities being ranked as low preference by multiple agents (children).
These are a few observations that can help the \prin{} to better cater to
the needs of the agents.

\noindent
\underline{\scmrs{} vs. \mcmrs{} in the \cc{} dataset.}
We recall
that the \cc{} dataset has two scenarios: with Zoom and chairs and without
these facilities. We note that there is not much difference in the
benefits. In particular, we can see that there is no significant decrease
in the improvement compared to the case when not having these facilities.
This seems to indicate that the university is well prepared to the COVID-19
special needs.

\noindent
\underline{Increase in benefit with budget.}
We observe that as the budget increases, in the case of \cc{}, the
probability of being matched increases gradually under both cost schemes. Also, the benefits for both cost schemes are comparable. However, in the
case of \pv{}, we observe an interesting threshold effect in the case of
\costi. For example, when~$p_\varnothing<0.33$, until a budget of~$4$, there is
no appreciable increase in the probability. However, for~$\beta=4$, the
probability is almost~$1$. We observed a similar phenomenon in the case of
\costii{} for budget~$6$, which is not presented in the plot.
This knowledge of the required budget can help us to give agents an indication of the budget needed to achieve a reasonable improvement in probability. 
\iftoggle{arxiv}{
In addition, further analysis of the activities that become available after the threshold is exceeded can give us an indication of whether we can modify some of these activities to make them available with a lower budget.
}{
}

\vspace*{-0.1in}
\section{Limitations and Future Work}
\label{sec:concl}
\vspace*{-0.05in}

A natural direction for future work is to extend
the framework to allow changes to the restrictions of multiple agents.
In such cases, an optimal allocation solution (e.g., Nash 
equilibrium~\cite{nash}) can be considered.
Our work assumes that each agent is matched to a single
resource. So, another direction is to extend the advice framework
by allowing agents to specify the number of resources needed.
In such a case, when an agent does not receive the requested number
of resources, the agent may be advised to either change her restrictions
or reduce the number of requested resources. We note that our framework 
can be extended to many scenarios where resources can be shared.  In such cases, for a shared resource, one can simply create
copies of resources with identical properties.

%
%
%

\medskip

\noindent
\textbf{Acknowledgments:}~ We are grateful to the reviewers of EUMAS~2022 for carefully reading the manuscript and providing valuable suggestions.
This work was supported by
Israel Science Foundation under grant 1958/20, the
EU Project TAILOR under grant 952215, 
Agricultural AI for Transforming Workforce and Decision Support (AgAID) grant no. 2021-67021-35344 from the USDA National Institute of Food and Agriculture, and the US
National Science Foundation grant OAC-1916805 (CINES).


\bibliographystyle{splncs04}

\iftoggle{arxiv}{
\clearpage

\appendix

\bigskip\bigskip

\begin{center}
\fbox{{\Large\textbf{Appendix}}}
\end{center}

\section{Additional Material for 
Section~\ref{sec:preliminaries}}
\label{sup:sec:preliminaries}

\noindent
\textbf{I.~Statement and Proof of Lemma~\ref{lem:matching_size}}

\medskip

\noindent
\textbf{Statement of 
Lemma~\ref{lem:matching_size}.} 
Let~$G$ denote the original compatibility graph and~$G'\ne G$
denote the compatibility graph obtained after some restrictions of
agent~$\sag$ are removed.
\begin{enumerate}[leftmargin=*,noitemsep,topsep=0pt]
\item Any maximum matching in $G'$ that was not 
a maximum matching in $G$ matches agent~$\sag$ to a new resource.
In addition, the edge from~$\sag$ to the matched
resource is not in $G$.
\item The size of a maximum matching in $G'$
is at most one more than that of $G$.
\end{enumerate}
\smallskip

\noindent
\textbf{Proof.}~
Since~$G'$ is obtained by adding new edges between~$\sag$ and one or more
resources that were previously incompatible, every new edge added to $G$ to
produce $G'$ must be incident with~$\sag$.  It can be seen that every
maximum matching in~$G'$ that is not a maximum matching in~$G$ must use one
of the new edges added to $G$.  Since each such edge is incident
with~$\sag$, the statement of Part~1 holds.

Let~$\ell$ and~$\ell'$ denote the sizes of maximum matchings in~$G$
and~$G'$ respectively. For the sake of contradiction, suppose $\ell' \geq
\ell+2$. Then, there is a maximum matching~$M'$ in~$G'$ that is not a
maximum matching in~$G$.  By Part~1, $M'$ must contain an edge $\{\sag,
y_j\}$. Let $M$ denote the matching obtained by deleting the edge~$\{\sag,
y_j\}$ from~$M'$. Since every edge in~$M$ is also an edge in $G$, it
implies that~$M$ is a matching in~$G$ with at least~$\ell+1$ edges, a
contradiction to the assumption that the maximum matching size in~$G$
is~$\ell$.
\QED

\medskip\medskip


\section{Additional Material for 
Section~\ref{sec:hardness}}
\label{sup:sec:hardness}

\smallskip

\textbf{I.~ Statement and Proof of 
Theorem~\ref{thm:or_hardness}}

\smallskip

\noindent
\textbf{Statement of 
Theorem~\ref{thm:or_hardness}.} \tdproblem{} is \cnp-hard for the
\mcsrs{} advice framework.


\medskip

\noindent
\textbf{Proof:}~ 
Our proof uses a reduction from the \maxcover{} problem
which is defined as follows: given a universal set
$U = \{u_1, u_2, \ldots, u_r\}$,
a family $F = \{F_1, F_2, \ldots, F_m\}$, where $F_j \subseteq U$,
$1 \leq j \leq m$, and integers $q$ and $t$, is there a subfamily $F'$ of $F$
such that $|F'| \leq q$ and the union of the sets in $F'$ has
at least $t$ elements?
The \maxcover{} problem is known to be \cnp-complete \cite{GareyJohnson79}.

Let $I$ denote an instance of \maxcover{} specified using the
parameters $U$, $F$, $q$ and $t$.
From $I$, we construct an instance $I'$ of \tdproblem{}
as follows.

\begin{enumerate}[leftmargin=*,noitemsep,topsep=0pt]
\item The agent set \agset{} = $\{\sag, x_1, x_2, \ldots, x_{r}\}$ has
$r+1$ agents while the resource set \rset{} =
$\{y_1, y_2, \ldots, y_{r}\}$ has $r$ resources.
(Recall that $r$ = $|U|$.)
The set \rset{} is in one-to-one correspondence with $U$ and
$\sag$ is the agent seeking advice.

\item Initially, the compatibility graph $G(\agset, \rset, E)$ has the $r$
edges given by $\{x_i, y_i\}$, $1 \leq i \leq r$. These edges form
a maximum matching of size $r$ in $G$.
(Thus, the number of maximum matchings in $G$ is 1.)

\item Initially, no edge is incident on $\sag$;
in other words, the probability that a maximum matching in $G$
includes $\sag$ is zero.

\item The set $R = \{R_1, R_2, \ldots, R_m\}$ associated
with $\sag$ is in one-to-one correspondence with the set
collection $F = \{F_1, F_2, \ldots, F_m\}$.

\item Suppose $F_i = \{u_{i_1}, u_{i_2}, \ldots, u_{i_{\ell}}\}$, where
$\ell = |F_i|$.
Corresponding to $F_i$, we have the following set of
resource-restriction pairs: $\{(y_{i_j}, \{R_i\}) ~:~ 1 \leq j \leq \ell\}$.
Thus, removing restriction $R_i$ adds the following set of $\ell$ edges
$\{\{\sag, y_{i_j}\}~:~ 1 \leq j \leq \ell\}$ to the initial
compatibility graph $G$.
Thus, we can think of removing restriction $R_i$ as adding
the edges from $\sag$ to the nodes in \rset{} corresponding
to the elements of $F_i$. Since each of these edges has
exactly one restriction, namely $R_i$, this corresponds to
the \mcsrs{} case.
Note that when these edges are added to $G$,
the size of the maximum matching remains $r$.
However, the number of maximum matchings increases to
$|F_i|+1$, and $|F_i|$ of these maximum matchings include $\sag$.
Thus, the increase in probability due to removing restriction $R_i$ is
$|F_i|/(|F_i|+1)$, $1 \leq i \leq m$.

\item The cost of removing each restriction $R_i$
is set to 1.
The budget $\beta$ is set to $q$, the budget on the number of sets
in the \maxcover{} instance.
The required increase in probability for $\sag$ is set to
$t/(t+1)$, where $t$ is the
coverage requirement in the \maxcover{} instance.
\end{enumerate}
This completes the construction of the instance $I'$ of \tdproblem.
It is easy to see that the construction can be carried out
in polynomial time.
We now show that there is a solution to the \tdproblem{} instance $I'$
iff there is a solution to the \maxcover{} instance $I$.

\noindent
\underline{If Part:}~ Suppose the \maxcover{} instance $I$ has a solution.
Without loss of generality, we can assume that the solution $F'$ is given by
$F' = \{F_1, F_2, \ldots, F_q\}$ and that the sets in $F'$
cover $t' \geq t$ elements of $U$.
For the \tdproblem{} instance, we choose the set $R'$ of restrictions
to be removed as $\{R_1, R_2, \ldots, R_q\}$.
which is obtained by choosing the
restrictions corresponding to the sets in $F'$.
The cost of adding these variables is $q = \beta$.
Since the sets in $F'$ cover $t' \geq t$ elements of $U$,
by our construction, the removal of the restrictions in $R'$
adds $t'$ edges between $\sag$ and the resources.
However, the maximum matching in the resulting graph remains $r$.
Thus, in the new compatibility graph
the number of maximum matchings of size $r$ is $t'+1$ and $t'$
of these matchings contain $\sag$.
Thus, the probability of $\sag$ being matched
in the new compatibility graph is at least $t'/(t'+1)$.
Since $t' \geq t$, it can be seen that the increase
in probability is at least $t/(t+1)$.
Thus, the \tdproblem{} instance $I'$ has a solution.

\noindent
\underline{Only If Part:}~ Suppose the \tdproblem{} instance $I'$ has a solution.
Without loss of generality, we may assume that the set $R'$ of
restrictions that are removed is
$R' = \{R_1, R_2, \ldots, R_q\}$.
Further, the increase in probability of $\sag$ due to this solution
is at least $t/(t+1)$.
By our construction, it can be seen that the removal of $R'$
must add at least $t$ edges between $\sag$ and the resources.
Now, consider the subfamily $F' = \{F_1, F_2, \ldots, F_q\}$.
The removal of each restriction $R_i$ in $R'$ adds the edges corresponding
to the elements in $F_i$, $1 \leq i \leq q$.
Since the removal of all the restrictions in $R'$
causes at least $t$ edges to be added to $\sag$,
it follows that the union of the sets in $F'$ covers at least
$t$ elements of $U$.
Thus, $F'$ is a solution to the \maxcover{} instance $I$, and
this completes our proof of Part~(a) of
Theorem~\ref{thm:or_hardness}. \QED

\section{Additional Material for 
Section~\ref{sec:alg_notation}}
\label{sup:sec:alg_notation}

\smallskip

\noindent
\textbf{I.~ Definitions of Monotone, Submodular and Supermodular\newline Functions:}
\begin{definition}\label{def:modular}
Let $X$ be a set and let $\mathbb{N}$ denote
the set of non-negative integers.
Suppose $f \::\: 2^{X} \rightarrow \mathbb{N}$ is
a function.
\begin{description}
\item{(a)} Function $f$ is \textbf{monotone non-decreasing} if for any
$P$ and $Q$ such that $P \subseteq Q \subseteq X$,
$f(P) ~\leq~ f(Q)$.

\item{(b)} Function $f$ is \textbf{submodular} if for any
$P$ and $Q$ such that $P, Q \subseteq X$,
$f(P) + f(Q)  ~\geq~ f(P \cup Q) + f(P \cap Q)$.

\item{(c)} Function $f$ is \textbf{supermodular} if for any
$P$ and $Q$ such that $P, Q \subseteq X$,
$f(P) + f(Q)  ~\leq~ f(P \cup Q) + f(P \cap Q)$.

\item{(d)} Function $f$ is \textbf{modular} if is both
submodular and supermodular.
\end{description}
\end{definition}

\medskip

\section{Additional Material for 
Section~\ref{sec:dm_scenario_id}}
\label{sup:sec:dm_scenario_id}

\smallskip

\noindent
\textbf{I.~ Statement and Proof of Lemma~\ref{lem:sc1}:}

\smallskip

\noindent
\textbf{Statement of Lemma~\ref{lem:sc1}:}~
Let~$M$ be a maximum matching and~$x\in \agset\,\cap\,\dme$. Adding
edge~$\{x,y\}$, where $y\in\rset$ is an incompatible resource,
increases the matching size iff ~$y\in \dme$.

\smallskip

\noindent
\textbf{Proof:}~
We will use the following notation: For two nodes~$u$ and~$v$
participating in a path~$P$, let~$uPv$ denote the subpath from~$u$
to~$v$ in~$P$. We have three cases: (i)~$y\in\dme$, (ii)~$y\in\dmu$,
and (iii)~$y\in\dmo$.

\smallskip

\noindent
\textbf{Case 1: $y\in\dme$.} We will show that the matching size increases if 
the edge~$\{x,y\}$ is added. Let~$P_x$ and~$P_y$ denote even-length alternating
paths from~$x$ to a free node~$f_x$ and from~$y$ to a free node~$f_y$
respectively. Note that if~$x$ is a free node, then~$P_x$ corresponds to a
path of length~$0$ containing just~$x$. The same holds for~$y$ as well. 
We will show that~$P_x$ and~$P_y$ are disjoint. Let~$v$ be a node common to 
both~$P_x$ and~$P_y$. If~$v\in X$, then,~$vP_yf_y$ is an odd-length alternating
path, while~$vP_xf_x$ is an even-length alternating path contradicting the
fact that~$\dme$ and~$\dmo$ are disjoint (Lemma~\ref{lem:dm}(1)).
Similarly, if~$v\in Y$, then,~$vP_xf_x$ is an odd-length alternating path,
while~$vP_yf_y$ is an even-length alternating path, again a contradiction.
Therefore,~$P_x$ and~$P_y$ are disjoint and therefore, the
path~$f_xP_xxyP_yf_y$ is an augmenting path in the new bipartite graph.

\smallskip
\noindent
\textbf{Case 2: $y\in\dmu$.} 
We will show that the matching size does not increase when edge~$\{x,y\}$
is added. Note that if adding this edge increases the matching size, then,
there exists an augmenting path~$P$ with respect to the matching~$M$.
Since~$M$ is a maximum matching in the original graph,~$P$ has to
necessarily contain the new edge. Also, since~$P$ is an augmenting path,
both its end points are free nodes. Therefore,~$P$ is of the
form~$P=f_xP'xyP''f_y$ where~$f_x$ and~$f_y$ are free nodes. If~$y\ne f_y$,
this implies that either~$y$ is reachable from free node~$f_y$ via the
alternating path~$yP''f_y$ in the original graph, a contradiction to the
fact that~$y\in\dmu$. Moreover, since~$y\in\dmu$,~$y$ cannot be a free
node, and therefore,~$y\ne f_y$. Hence,~$P$ cannot exist.

\smallskip
\noindent
\textbf{Case 3: $y\in\dmo$.} We will show that the matching size does not
increase when edge~$\{x,y\}$ is added. As in Case~2, suppose the matching
size increases, then there exists an augmenting path~$P$ of the
form~$P=f_xP'xyP''f_y$, where~$f_x$ and~$f_y$ are free nodes.
If~$y\in\dmo$, then,~$P'$ is an even-length path since~$f_xP'x$ is
odd-length,~$\{x,y\}$ is odd-length, and~$yP_yf_y$ is even-length. This
contradicts the fact that any augmenting path has to be odd-length.

\smallskip
Therefore, an augmenting path cannot be formed when the edge~$\{x,y\}$ is
added with~$y\in\dmu\cup\dmo$, and in turn, the matching size cannot
increase. \QED

\medskip

\section{Additional Material for Section~\ref{sec:disjunction}}
\label{sup:sec:disjunction}

\noindent
\textbf{I.~ Statement and Proof of Lemma~\ref{lem:or_submodular}}

\smallskip

\noindent
\textbf{Statement of Lemma~\ref{lem:or_submodular}:}
Consider a matching advice framework with the \mcsr{}
incompatibility. Then, for Matching Scenario~2, the new matching count
function~$f(\cdot)$ is monotone submodular.

\smallskip

\noindent
\textbf{Proof.}~
Since for any~$A$ and $A'$ such that $A\subseteq A'$, ~$G_A$ is a subgraph of~$G_{A'}$ and the maximum
matching size in~$G_{A'}$ is the same as that in~$G_A$, it follows that any new matching
in~$G_A$ is also a new matching in~$G_{A'}$. Therefore,~$f$ is monotone. Now, we
will show that for any~$B,B'\subseteq\restset$, $f(B)+f(B')\ge f(B\cup
B')+f(B\cap B')$.  Suppose~$y\in\sagrset$ is an incompatible resource and
let~$B_y\subseteq\restset$ be the set of restrictions such that for~$r\in
B_y$, removing~$r$ makes~$y$ compatible, or in other
words,~$(y,\{r\})\in\incset$ iff~$r\in B_y$. Resource~$y$ belongs to one of the
following four classes.
\vspace*{-0.07in}
\begin{align}\label{eqn:classes}
y\in\left\{
\begin{array}{ll}
C_1, & \text{if $B_y\cap (B\cap B')\ne \varnothing$}, \\ 
C_2, & \text{if $B_y\cap B\ne\varnothing$ and $B_y \cap B'=\varnothing$}, \\
C_3, & \text{if $B_y\cap B=\varnothing$ and $B_y \cap B'\ne\varnothing$}, \\
C_4, & \text{if $B_y\cap B\ne\varnothing$, $B_y \cap B'\ne\varnothing$ and} \\
& \text{$B_y\cap (B\cap B')=\varnothing$}\,.
\end{array}
\right.
\end{align}
\vspace*{-0.07in}

Let~$\omega_\ell$ denote the number of new maximum matchings obtained by
adding all edges~$\{\sag,y\}$,~$y\in C_\ell$ to~$G$. Note that~$f(B\cap
B')$ corresponds to maximum matchings obtained from relaxing all resources
belonging to~$C_1$, and therefore,~$f(B\cap B')=\omega_1$. The
quantity~$f(B)$ corresponds to maximum matchings resulting from adding
resources from~$C_1$,~$C_2$, and~$C_4$, and
therefore,~$f(B)=\omega_1+\omega_2+\omega_4$.
Similarly,~$f(B')=\omega_1+\omega_3+\omega_4$.
Finally,~$f(B\cup B')=\omega_1+\omega_2+\omega_3+\omega_4$.
Clearly,~$f(B)+f(B')\ge f(B\cup B')+f(B\cap B')$.
\QED

\medskip

\noindent
\textbf{II. Statement and Proof of Theorem~\ref{thm:or_greedy}:}

\smallskip

\noindent
\textbf{Statement of Theorem~\ref{thm:or_greedy}.}
Consider the \mcsrs{} incompatibility.
Suppose each restriction has the same removal cost  and the maximum
matchings of the compatibility graph~$G$ are chosen from the uniform
distribution. Then, given an oracle for computing the probability 
$p(\cdot)$, Algorithm~\ref{alg:or_greedy} provides a
solution to the \tproblem{} problem with cost at most~$\beta$ and benefit at 
least~$(1-1/e)$ of the optimal solution.

\medskip

\noindent
\textbf{Proof.}~Let~$G'$ denote the compatibility graph obtained after
removing the restrictions obtained by Algorithm~\ref{alg:or_greedy} and
$\Gopt$ denote the compatibility graph obtained by removing restrictions
from an optimal solution. Let~$\nonmatch$  and~$\oldmatch$ denote the
number of maximum matchings in the old compatibility graph where~$\sag$ is
not matched and matched respectively. Let~$\matchcurr$ and~$\matchopt$
denote the number of new maximum matchings in~$G'$ and~$\Gopt$ respectively. From Lemma~\ref{lem:matching_size}, it follows that
any new matching in the new compatibility graph ($G'$ or~$\Gopt$) must
have~$\sag$ matched. Hence, the total number of maximum matching in~$G'$
and~$\Gopt$ are~$\nonmatch+\oldmatch+\matchcurr$ and~$\nonmatch+\oldmatch+\matchopt$
respectively.

By Lemma~\ref{lem:or_submodular}, Algorithm~\ref{alg:or_greedy} provides a
solution for which the total number of new maximum matchings is at
least~$(1-1/e)$ times the total number of new maximum matchings in an optimal
solution. Therefore,~$\matchcurr\ge(1-1/e)\,\matchopt$.
Let~$\pcurr=\frac{\oldmatch+\matchcurr}{\nonmatch+\oldmatch+\matchcurr}$
and~$\popt=\frac{\oldmatch+\matchopt}{\nonmatch+\oldmatch+\matchopt}$ denote the probability
that~$\sag$ is matched in~$G'$ and~$\Gopt$ respectively.
\begin{align*}
    \frac{\pcurr}{\popt}&=\frac{\oldmatch+\matchcurr}{\oldmatch+\matchopt} 
    \frac{\nonmatch+\oldmatch+\matchopt}{\nonmatch+\oldmatch+\matchcurr} 
    \ge\frac{\oldmatch+\matchcurr}{\oldmatch+\matchopt}\\
    &\ge\frac{\oldmatch+(1-1/e)\matchopt}{\oldmatch+\matchopt}\\
    &=\frac{(\oldmatch+\matchopt)-(1/e)\matchopt}{\oldmatch+\matchopt}
    =1 - \Big(\frac{1}{e}\,\frac{\matchopt}{\oldmatch+\matchopt}\Big)\\
    &\ge\Big(1-\frac{1}{e}\Big)\,.
\end{align*}
This completes our proof of 
Theorem~\ref{thm:or_greedy}. \QED

\medskip\medskip

\section{Additional Material for Section~\ref{sec:threshold}}
\label{sup:sec:threshold}

\smallskip

\noindent
\textbf{I.~ Statement and Proof of Theorem~\ref{thm:threshold}:}

\smallskip

\noindent
\textbf{Statement of 
Theorem~\ref{thm:threshold}:}
For \tproblem{} with threshold-like incompatibility,
given an oracle for probability~$p(\cdot)$, Algorithm~\ref{alg:thresh} provides an optimal solution in  $O(\beta^\alpha\log|\restset|)$ calls to the probability~$p(\cdot)$ computing oracle, where~$\beta$ is the
budget,~$\restset$ is the restrictions set of special agent~$\sag$, and~$\alpha$ is the number of blocks in~$\rset$.

\smallskip

\noindent
\textbf{Proof.~}
First, we will show that the algorithm provides an optimal solution. 
Let~$A$ be any set of restrictions. We will show that there
exists an~$A'\subseteq A$ such that~$\cost(A')\le\cost(A)$
and~$\benefit(A')\ge\benefit(A)$ satisfying the following: for any
block~$\restset_\ell$,~$A'\cap\restset_\ell=\{r_{\ell,s},
r_{\ell,s+1},\ldots,r_{\ell,t(\ell)}\}$ for some~$1\le s\le t(\ell)$.  Let
$\coll = \{\restset'\mid (y,\restset')\in\incset$ and~$y$ becomes
compatible after relaxing~$A\}$. By definition, every~$\restset'$ is of the
form~$\{r_{\ell,s'}, r_{\ell,s'+1},\ldots,r_{\ell,t(\ell)}\}$. We simply
choose~$A'=\bigcup_{\restset'\in\coll}
\bigcup_{\ell=1}^\alpha\restset'\cap\restset_\ell$,~$1\le \ell\le\alpha$.
It is easy to see that~$A'$ is of the form described above
where~$r_{\ell,s}=\min_{\restset'\in\coll} \min \restset'$.
Since,~$\bigcup_{\restset'\cup\coll}\restset'\subseteq A$, it follows
that~$A'\subseteq A$. Therefore, any optimal solution must have the
property of~$A'$. 

Now consider any two sets~$A_1\subseteq A_2$ that satisfy the property
of~$A'$.  Since~$\benefit(A_2)\le\benefit(A_1)$, given a
budget~$\beta_\ell$ for block~$\ell$, the gain can be maximized by choosing
the least element possible from the set~$\restset_\ell$. This is being
performed in Line~\ref{alg:threshold:min} in the algorithm. This proves the
first part of the statement.

Now we will bound the time complexity of the algorithm. Firstly, we note
that the number of $\alpha$-partitions of~$\beta$ is $O(\beta^\alpha)$, and
can be enumerated in as much time. Secondly, given a budget~$\beta_\ell$
for clause~$\ell$, Line~\ref{alg:threshold:min} can be computed
in~$O(\log|\restset_\ell|)$ time (using a binary search given the natural
ordering of the variables in this set).
\QED

\section{Additional Material for Section~\ref{sec:experiments}}
\label{sup:sec:experiments}

\begin{figure}[hbt]
\centering
\includegraphics[width=.5\textwidth]{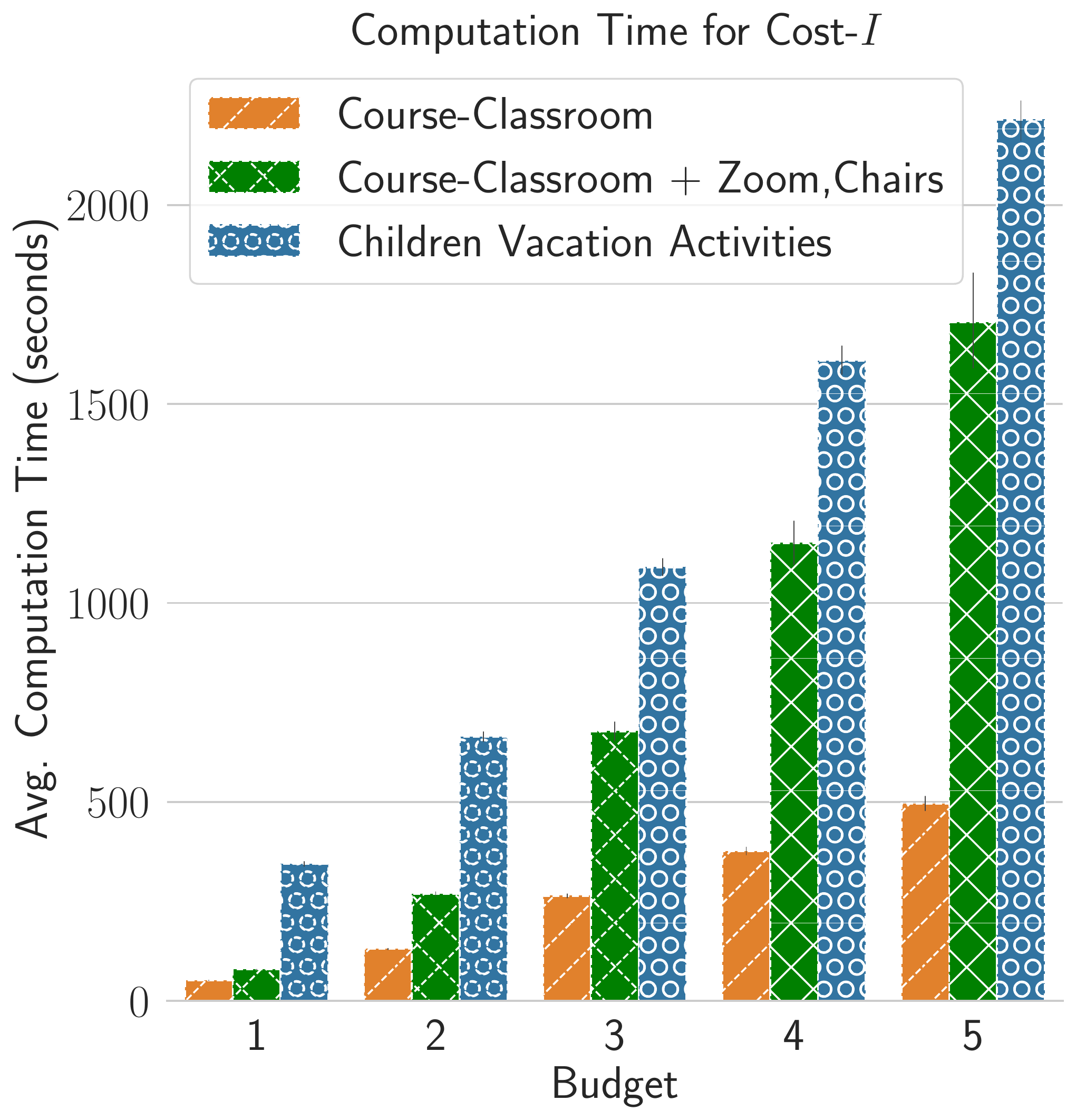}
\caption{Computation time for the different datasets and varying budget for \costi{}. Computation time for \costii{} was similar. The error bars represent
the size of 95\% confidence intervals.  \label{fig:comptime}}
\end{figure}
\noindent
\textbf{Computation time of Algorithm~\ref{alg:thresh}:}
The time required to run Algorithm~\ref{alg:thresh} is provided in the bar chart of
Figure~\ref{fig:comptime}.  The computation time for the \pv{} dataset is
much longer even for lower budget since the corresponding compatibility
graph is larger compared to the \cc{} dataset.  Also,  the addition
of attributes in the \cc{} dataset (Zoom and chairs) significantly increases
the computation time. The same holds as the budget is
increased, as there are many more partitions of the budget to evaluate.

}{}  
\end{document}